\documentclass{article}

\usepackage{arxiv}

\usepackage[utf8]{inputenc} 
\usepackage[T1]{fontenc}    
\usepackage{hyperref}       
\usepackage{url}            
\usepackage{booktabs}       
\usepackage{amsfonts}       
\usepackage{nicefrac}       
\usepackage{microtype}      
\usepackage{lipsum}		
\usepackage{graphicx}
\usepackage{natbib}
\usepackage{doi}
\usepackage{amsmath}
\usepackage{float}
\usepackage{subcaption}

\graphicspath{ {Figures/} }

\title{Coupled VAE: Improved Accuracy and Robustness of a Variational Autoencoder}

\author{ \href{https://orcid.org/0000-0001-8045-0404}{\includegraphics[scale=0.06]{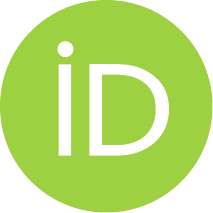}\hspace{1mm}Shichen Cao}\\
	Worcester Polytechnic Institute\\
	Worcester, MA USA 01609\\
	\texttt{cao.schen@gmail.com} \\
	\And
	\href{https://orcid.org/0000-0002-2331-7153}{\includegraphics[scale=0.06]{orcid.pdf}\hspace{1mm}Jingjing Li} \\
	Boston University\\
	Boston, MA USA 02215\\
	\texttt{jli0203@bu.edu} \\
		\And
	{\hspace{1mm}Kenric Nelson} \\
	Photrek\\
	Watertown, MA USA 02472\\
	\texttt{kenric.nelson@gmail.com} \\
		\And
	{\hspace{1mm}Mark Kon} \\
	Boston University\\
	Boston, MA USA 02215\\
	\texttt{mkon@bu.edu} \\
}

\date{}

\hypersetup{
pdfsubject={q-bio.NC, q-bio.QM},
pdfauthor={David S.~Hippocampus, Elias D.~Striatum},
pdfkeywords={First keyword, Second keyword, More},
}

\begin{document}
\maketitle

\begin{abstract}
	We present a coupled Variational Auto-Encoder (VAE) method, which improves accuracy and robustness in generation of handwritten numeral images, while also reducing divergence between the posterior and a prior latent distribution. The new method weighs outlier samples with a higher penalty by generalizing the original loss function to a coupled entropy function, using the principles of nonlinear statistical coupling. 
We evaluate the performance of the coupled VAE model using the Modified National Institute of Standards and Technology (MNIST) dataset. Compared with the traditional VAE algorithm, the output images generated by the coupled VAE method in 20 dimensions are clearer and more distinct. Assessment of the posterior latent neural layer is completed with a 2D example, including a scatter-plot visualization and analysis of mean, standard deviation, and divergence.
This shows the posterior means maintain a tight, symmetric distribution around the prior mean of zero. We analyze the histogram of the likelihoods of the generated images using the generalized mean, which measures the algorithms’s accuracy as a function of the relative risk. This approach improves the neutral accuracy, formed as the geometric mean, which is consistent with measure of the Shannon cross-entropy. 
Such an improvement also occurs for the robust accuracy, measured by the -2/3 generalized mean.
\end{abstract}


\section{Introduction}
An overarching  challenge for machine learning is the development of methodologies that assure the accuracy and robustness of models given limited training data. The variational autoencoder (VAE) contributes toward this goal by learning a low-dimensional statistical model of input data, a model referred to as the latent (neural) layer.  Such a low dimensional latent model is optimized via variational inference methods, and consists of minimizing a loss function, the so-called negative evidence lower bound, which has two components. The first is a cross-entropy between the generated and the source data, while the second is a divergence between the prior and the posterior distributions of the latent layer model.

In this paper we show that accuracy and robustness can be improved by utilizing a generalization of the cross-entropy and divergence. This generalization of the entropy functions is referred to as the coupled entropy because it models deviation from additive Shannon entropy associated with the nonlinearity of a complex system. In the context of generative machine learning algorithms, the coupled entropy provides a method to optimize for robustness against outlier samples which may not be fully represented in a training set. Whereas the 
entropy measures the average uncertainty of a distribution with equal weighting of each training sample, the coupled-entropy adds/subtracts additional weight to samples in the tails of the distribution based on 
positive/negative coupling, respectively. The use of positive coupling in the cross-entropy 
and divergence costs of the variational autoencoder enables the learning of a robust inference model.

Our study builds from the work of \citet{Kingma2014} on variational autoencoders and  \citet{Tran2017} 
on deep probabilistic programming. Variational autoencoders are a unsupervised learning method for training encoder and decoder neural networks. Between the encoder and decoder, the parameters 
of a multidimensional distribution are learned to form a compressed latent representation of 
the training data \citep{Samuel2015}. It is an effective method for generating complex datasets 
such as images and speech. \citet{Zalger2017} implemented the application of VAE for aircraft turbomachinery design and \citet{Xu2018} used VAEs to achieve unsupervised anomaly detection for seasonal KPIs (key performance indicators) in web applications. Autoencoders
can use a variety of latent variable models, but restricting the models can enhance performance. 
Sparse autoencoders add a penalty for the number of active hidden layer nodes used in the model. Variational autoencoders further restrict the model to a probability distribution ${q_\phi }\left( {{\bf{z}}|{\bf{x}}} \right)$  
specified by a set of encoder parameters $\phi$ which approximates the actual conditional probability ${p}\left( {{\bf{z}}|{\bf{x}}} \right)$. Variational inference as reviewed by \citet{Blei2016}, is used to learn this approximation by minimizing an objective function such as Kullback-Liebler divergence. The decoder learns a set of parameters $\theta$ for a generative distribution ${q_\theta }\left( {\bf{x'}}|{{\bf{z}}} \right)$, where $\bf{z}$ is the latent variable, and $\bf{x'}$  is the output generated data. The complexity of the data distribution $p(\bf{x})$ makes direct computation of the divergence between the approximate and exact latent conditional probabilities intractable; however, a variational or evidence lower bound (ELBO) is computable and consists of two components, the expected reconstruction
log-likelihood of the generated data (cross-entropy) and the negative of the divergence between the latent posterior conditional probability ${q_\phi }\left( {{\bf{z}}|{\bf{x}}} \right)$ and a latent prior distribution $p(\bf{z})$, which is typically a standard normal distribution but can more sophisticated for particular model requirements.

Recently, \citet{higgins2016beta} proposed a $\beta$-VAE framework, which can provide a more disentangled latent representation $\bf{z}$ \citep{burgess2018understanding} by increasing the weight of the KL divergence term of the ELBO. Since the KL-divergence is a regularisation that constrains the capacity of the latent
information channel $\bf{z}$, increasing the weight of the regularization with $\beta > 1$ puts  pressure on the learnt posterior to be more tightly packed. The effect seems to be to encouragement of each dimension to store distinct information and excess dimensions as highly packed noise.

In this study, we draw upon the principles of Nonlinear Statistical Coupling (NSC) \citep{Nelson2010, Nelson2017}
to train and evaluate the accuracy and robustness of a variational autoencoder \citep{Nelson2020}.  
NSC is derived from non-extensive statistical mechanics \citep{Tsallis2009},  
and focuses on the role of nonlinear coupling $\kappa$ in generalizing entropy and its related functions. Robustness of autoencoders to outliers is critical for generating a reliable representation of particular data types in the encoded space when using corrupted training data \citep{akrami2019robust}. The approach defines a family of heavy-tailed (positive coupling) and compactly-supported (negative coupling) distributions which maximize a generalized entropy function referred to as coupled entropy. Using the MNIST dataset of handwritten numerals, we show that the measure of robustness and accuracy based on generalized information theory is improved by incorporating the coupled entropy into the loss 
function of the variational autoencoder.

The next two sections provide a description of the variational autoencoder and the MNIST dataset used for our evaluation. Section 4 introduces the generalized metrics which are used to measure the robustness and accuracy. In Section 5, the improved autoencoder is evaluated using the MNIST handwritten numeral test set. Section 6 discusses the results from using a simplified 2-dimensional latent variable, and Section 7 describes our conclusions.

\section{The Variational Autoencoder}
\begin{figure}
\centering
\includegraphics[width=5.5 cm]{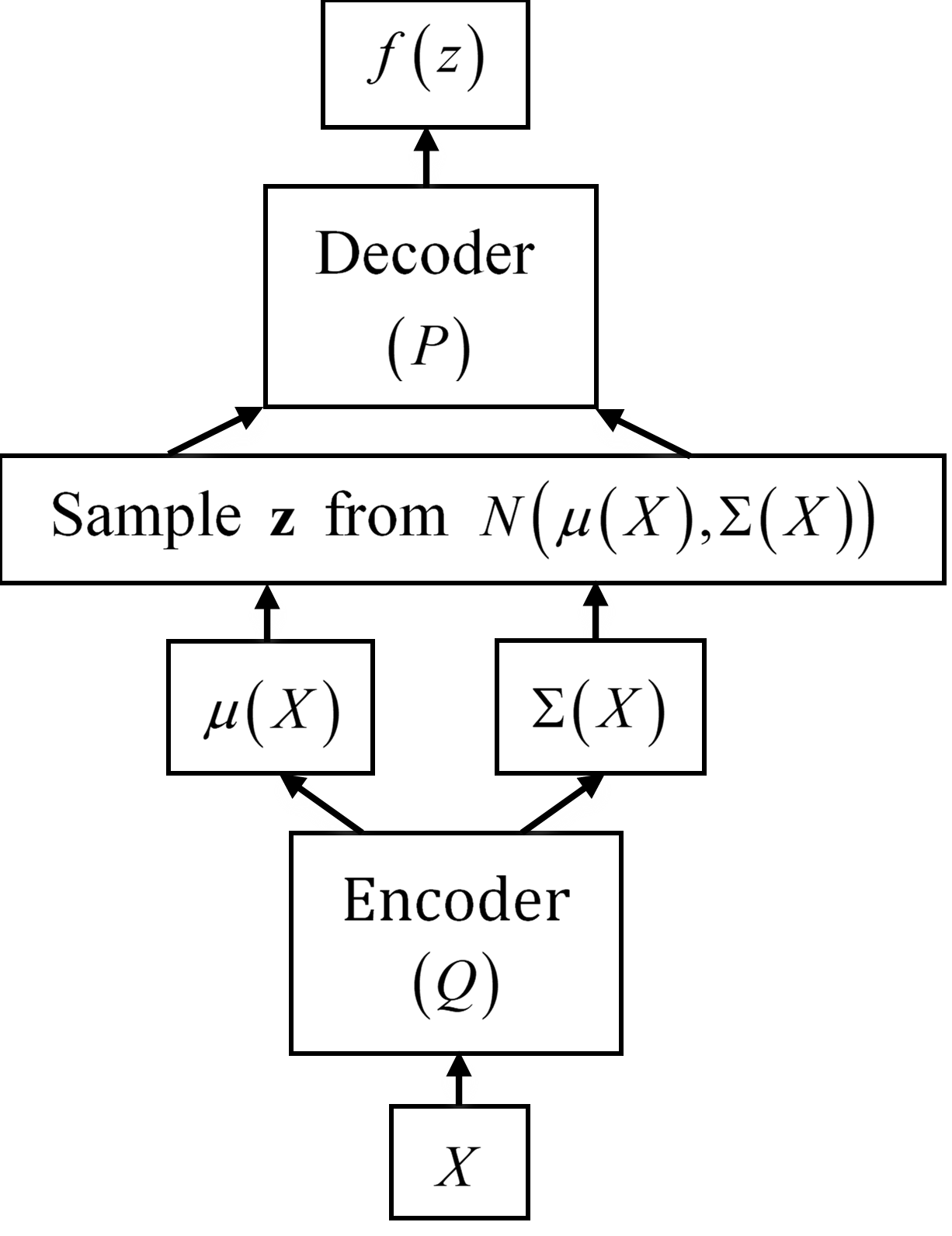}
\caption{The variational autoencoder consists of an encoder, a probability model and a decoder.\label{fig1}}
\end{figure}

A variational autoencoder consists of an encoder, a decoder, and a loss function. The encoder is a neural network that converts high-dimensional information from the input data into a low-dimensional hidden, 
latent representation $\bf{z}$. While in general autoencoders can learn a variety of representations, 
VAEs especially learn the parameters of a probability distribution. The model used here learns the means and standard deviations $\theta$ of a collection of multivariate Gaussian distribution and stores this information in a two-layered space. The training loss function, which is the negative of an evidence lower bound, is designed to optimize the approximation of this model's parameters to the actual conditional probability of the latent representation given the data. Figure \ref{fig1} represents the basic structure of an autoencoder. 

\subsection{VAE loss function}
The encoder reads the input data and compresses and transforms it into a fixed-shape latent representation $\bf z$, while the decoder decompresses and reconstructs the information from this latent representation, outputting specific distribution parameters to generate a new reconstruction $\bf{x'}$. The true posterior distribution $p({\bf{z}}|{\bf{x}}^{(i)})$ of $\bf{z}$ given ${\bf{x}}^{(i)}$ is unknown, but we use the Gaussian approximation $q({\bf{z}}|{\bf{x}}^{(i)})$ with mean vector $\mu^{(i)}$ and covariance matrix ${\rm diag}(\sigma_{1}^{2},\cdots,\sigma_{d}^{2})^{(i)}$ instead. The goal of the algorithm is to maximize the variational or evidence lower bound (ELBO) on the marginal density of individual datapoints. 


For a dataset ${\bf{X}} = \left\{{\bf{x}}^{(i)}\right\}_{i=1}^{N}$  consisting of $N$ independent and identically distributed samples, the variational lower bound for the $i^{th}$ data point or image ${\bf{x}}^{(i)}$ in the original VAE algorithm \citep{Kingma2014} is

\begin{equation}
ELBO\left( {{{\bf{x}}^{\left( i \right)}}} \right) =  - {D_{KL}}\left( {q\left( {{\bf{z}}|{{\bf{x}}^{\left( i \right)}}} \right)\parallel p\left( {\bf{z}} \right)} \right) + {\mathbb{E}_{q\left( {{\bf{z}}|{{\bf{x}}^{\left( i \right)}}} \right)}}\left[ {\log p\left( {{{\bf{x}}^{\left( i \right)}}|{\bf{z}}} \right)} \right].
\end{equation}

The first term on the right-hand side is the negative Kullback-Leibler divergence between the posterior variational approximation $q({\bf{z}}|{\bf{x}})$ and a prior distribution $\bf{z}$ which is selected to be a standard Gaussian distribution. The second term on the right-hand side is denoted as the expected reconstruction log-likelihood, and is  referred to as the cross-entropy. 
 Let $J$ be the dimensionality of $\bf{z}$; then the Kullback-Leibler divergence simplifies to

\begin{align}
    - {D_{KL}}\left( {q\left( {{\bf{z}}|{{\bf{x}}^{(i)}}} \right)||p\left( {\bf{z}} \right)} \right)=&\int_{ - \infty }^\infty  {q\left( {{\bf{z}}|{{\bf{x}}^{(i)}}} \right)\left( {\log p\left( {\bf{z}} \right) - \log q\left( {{\bf{z}}|{{\bf{x}}^{(i)}}} \right)} \right)} dz\\ =  &\frac{1}{2}\sum\limits_{j = 1}^J {\left( {1 + \log \left( {{{\left( {{\sigma _j}} \right)}^2}} \right) - {{\left( {{\mu _j}} \right)}^2} - {{\left( {{\sigma _j}} \right)}^2}} \right)}.\\
\end{align}

The expected reconstruction log-likelihood (cross-entropy) ${E_{q\left( {{\bf{z}}|{{\bf{x}}^{(i)}}} \right)}}\left[ {\log p\left( {{{\bf{x}}^{(i)}}|{\bf{z}}} \right)} \right]$ can be estimated by sampling, i.e.,

\begin{equation}
{\mathbb{E}_{q\left( {{\bf{z}}|{{\bf{x}}^{(i)}}} \right)}}\left[ {\log p\left( {{{\bf{x}}^{(i)}}|{\bf{z}}} \right)} \right] = \frac{1}{L}\sum\limits_{l = 1}^L {\left( {\log p\left( {{{\bf{x}}^{(i)}}|{{\bf{z}}^{(i,l)}}} \right)} \right)}.
\end{equation}

And if data $\bf{x}$ given $\bf{z}$ follow a multivariate Bernoulli distribution in dimension D, we have

\begin{equation}
\log p\left( {{\bf{x}}|{\bf{z}}} \right) = \sum\limits_{i = 1}^D ({{x_i}\log {y_i} + \left( {1 - {x_i}} \right)\log \left( {1 - {y_i}} \right)}),
\end{equation}
where $y$ is the output of the decoder using the Bernoulli distribution. Therefore, the regular loss function can be calculated by

\begin{equation}
L\left( {{{\bf{x}}^{\left( i \right)}}} \right) = -ELBO\left( {{{\bf{x}}^{\left( i \right)}}} \right) =  {D_{KL}}\left( {q\left( {{\bf{z}}|{{\bf{x}}^{\left( i \right)}}} \right)\parallel p\left( {\bf{z}} \right)} \right) - \frac{1}{L}\sum\limits_{l = 1}^L {\left( {\log p\left( {{{\bf{x}}^{\left( i \right)}}|{{\bf{z}}^{\left( {i,l} \right)}}} \right)} \right)}.
\end{equation}

For our work, the loss function is modified to improve the robustness of the variational autoencoder, something that will be discussed in Section 4.

\subsection{Comparison with other generative machine learning methods}
The paradigm of generative adversarial networks (GANs) is a recent advance in generative machine learning methods. The basic idea of GANs was published in a 2010 blog post by \citet{Olli2010}, and the name ‘GAN’ was introduced by \citet{Goodfellow2014}. In comparison with variational autoencoders, generative adversarial networks are used for optimizing generative tasks specifically. GANs can produce models with true latent spaces, as is the case of Bidirectional GAN (BiGAN) and adversarially learned inference (ALI) \citep{Donahue2017, Dumoulin2017}, which are designed to improve the performance of GANs, GANs.  However, GANs cannot generate reasonable results when data are high-dimensional \citep{neyshabur2017stabilizing}. By contrast, as a probabilistic model, the specific goal of a variational autoencoder is to marginalize out non-informative variables during the training process. The ability to use complex priors in the latent space enables existing expert knowledge to be incorporated.

Bayesian networks form another generative model. Judea Pearl proposed the Bayesian network paradigm in 1985. Bayesian networks have a strong ability to capture the symbolic figures of input information \citep{Pearl1985} and combine objective probabilities with subjective estimates for both qualitative and quantitative modeling. The basic  concept of Bayesian networks is built on Bayes’ theorem. 
Another effective way to solve for the posterior of the distribution derived from neural networks is to train and predict using variational inference techniques \citep{Goodfellow2016}. Compared with the original Bayesian network, the basic building blocks of deep networks provide multiple loss functions for making multi-target predictions, for transfer learning, and for varying outputs to depending on situations. The improvement of the deeper architectures, using VAE specifically, continues to  grow.

Other generative models are now commonly combined with a variational autoencoder to improve performance. \citet{Ebbers2017} developed a VAE with a Hidden Markov Model (HMM) as the latent model for discovering acoustic units. \citet{Dilokthanakul2016} studied the use of Gaussian mixture models as the prior distribution of the VAE to perform unsupervised clustering through deep generative models. He showed a heuristic algorithm called “minimum information constraint,” and it is capable of improving the unsupervised clustering performance with his model. \citet{Srivastava2017} presented the effective autoencoding variational Bayes based inference method for latent Dirichlet allocation (LDA). This model solves the problems caused by autoencoding variational Bayes by the Dirichlet prior and by component collapsing. Also, this model matches traditional methods inaccuracy with much better inference time.

\section{Use of MNIST database for evaluation}
The MNIST, handwritten digit database is a large database of handwritten digits consisting of a training set of 60,000 images and a test set of 10,000 images widely used for evaluating machine learning and pattern recognition methods. The digits have been
size-normalized and centered in a fixed-size image. Each image in the database contains 28 by 28 grey-scale pixels. Pixel values vary from 0 to 255. Zero means the pixel is white, or background, while 255 means the pixel is black, or foreground \citep{Yann}.

For this research, we used the MNIST database as the input. Specifically, input x is a batch of the 28 by 28-pixel photo of a handwritten number. The encoder encodes the data, which is 784-dimensional for each image in a batch into the latent layer space z. For our experiment, the dimension of space z can be chosen from 2 to 20. Taking the latent layers z as the input, the probability distribution of each pixel is computed using a Bernoulli or Gaussian distribution by the decoder. The decoder outputs corresponding 784 parameters and decodes the remodeled value to generate the images at the last step. We used specific numbers of images from the training set as the batch size and fixed epochs for the most modeling process. Additionally, in the learned MNIST manifold, visualizations of learned data and reproduced results can be plotted in the research.

\section{Accounting For Risk with Coupled-entropy}
Machine learning algorithms, including the VAE, have achieved efficient learning and inference for many image processing applications. Nevertheless, assuring accurate forecasts of the uncertainty is still a challenge.  Problems such as outliers and overfitting impact the robustness of scientific prediction and engineering systems. This paper concentrates on assessing and improving the robustness of the VAE algorithm.
\subsection{Assessing probabilistic forecasts with the generalized mean}
First, proper metrics are needed to evaluate the accuracy and robustness of machine learning algorithms, such as VAE. The arithmetic mean and the standard deviations are widely used to measure central tendency and fluctuation, respectively, of a random variable. Nevertheless, a random variable formed by the ratio of two independent random variables has a central tendency determined by the geometric mean, as described by \citet{Donald1879}. Thus, probabilities which are formed as ratios need the geometric mean to measure the central tendency.

A Risk Profile, which is the spectrum of the generalized means of probabilities, was introduced to evaluate the central tendency and fluctuations of probabilistic inferences. The generalized mean $(\frac{1}{N}\sum\limits_{i=1}^{N} p_{i}^{r})^{\frac{1}{r}}$ is a translation of generalized information-theoretic metrics back to the probability domain, and is derived in the next section. It's use as a metric for evaluating and training inference algorithms is related to the Wasserstein distance \citep{Frogner2015}, which incorporates the generalized mean. The accuracy of the likelihoods is measured with robust, neutral, and decisive risk bias using the  $r = -\frac{2}{3}$, $r = 0$ (geometric), and  $r = 1$ (arithmetic) means, respectively. With no risk bias ($r = 0$) the geometric mean is equivalent to transforming the cross-entropy between the forecast $p_{i}$ and the distribution of the test samples to the probability domain. The arithmetic mean ($r = 1$) is  a simple measure of the decisiveness (i.e. were the class probabilities in the right order so that a correct decision can be made?). This measure de-weights probabilities near zero since increasing $r$ reduces the influence of small probabilities on the average. To complement the arithmetic mean we choose a negative conjugate value. The conjugate is not the harmonic mean ($r = -1$) because this turns out to be too severe a test. Instead,  $r = -\frac{2}{3}$ is chosen based on a dual transformation between heavy-tail and compact-support domains of the Coupled Gaussian distribution. The risk sensitivity $r$ can be decomposed into the nonlinear coupling and the power and
dimension of the variable $r(\kappa,\alpha,d) = \frac{-\alpha \kappa}{1+d\kappa}$. With the alpha term dropping out, the dual transformation has the following relationship $\hat{\kappa} \Leftrightarrow \frac{-\kappa}{1+d\kappa}$. Taking $\alpha = 2$ and $d =1$ , the coupling for a risk bias of one is $1 = \frac{-2\kappa}{1+\kappa} \Rightarrow \kappa = -\frac{1}{3}$ and the conjugate values are $\hat{\kappa} = \frac{\frac{1}{3}}{1-\frac{1}{3}} = \frac{1}{2}$ and $\hat{r} = \frac{-2\cdot \frac{1}{2}}{1+\frac{1}{2}} = -\frac{2}{3}$ \citep{Nelson2020}. The Robustness metric increases the weight of probabilities near zero since negative powers invert the probabilities prior to the average.

For simplicity, we refer to these three metrics as the robustness, accuracy, and decisiveness.  The label “accuracy” is used for the neutral accuracy, since “neutralness” is not appropriate and “neutral” does not express that this metric is the central tendency of the accuracy. Summarizing:

\begin{equation}
Decisiveness{\,\, \rm{  (Arithmetic \,Mean): \,\,}} \frac{1}{N}\sum\limits_{i = 1}^N {{p_i}} .
\end{equation}
\begin{equation}\label{geomean}
Accuracy{\,\,\rm{  (Geometric \,Mean): \,\,}} \prod\limits_{i = 1}^N {p_{i}^{\frac{1}{N}}}.
\end{equation}
\begin{equation}
Robustness{\,\,\rm{  ( - 2/3 \, Mean):\,\, }} {\left( \frac{1}{N}\sum\limits_{i = 1}^N {p_{i}^{-\frac{2}{3}}} \right)^{-\frac{3}{2}}}.
\end{equation}

\begin{figure}
\centering
        \begin{subfigure}[b]{0.3\textwidth}
                \centering
                \includegraphics[width=.85\linewidth]{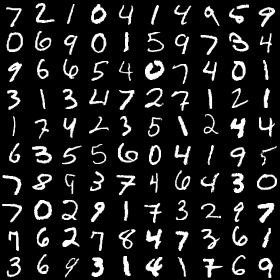}
                \caption{MNIST input images}
        \end{subfigure}%
                \begin{subfigure}[b]{0.3\textwidth}
                \centering
                \includegraphics[width=.85\linewidth]{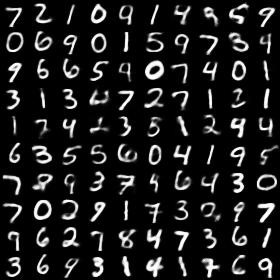}
                \caption{VAE generated output}
        \end{subfigure}%
\caption{Example set of (\textbf{a}) MNIST input images and (\textbf{b}) VAE generated output images.\label{fig2}}
\end{figure}

\begin{figure}
    \centering
    \includegraphics[width=0.6\textwidth]{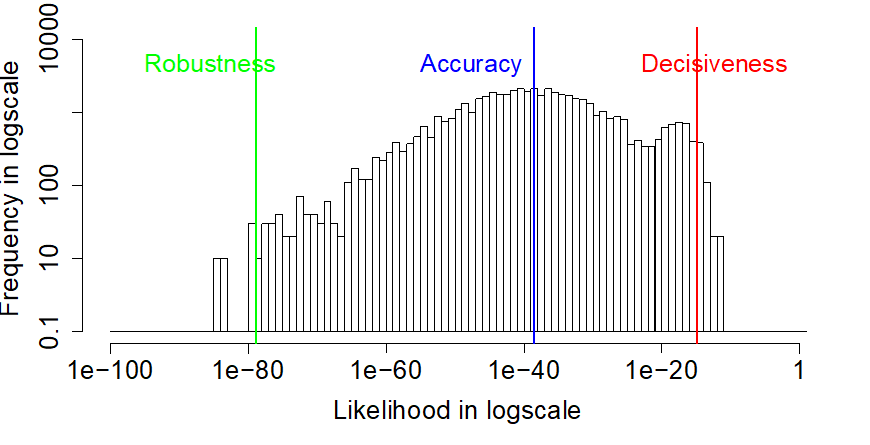}
\caption{The likelihood for the input images under the VAE model. The extremely small value of -2/3 mean metric indicates the poor robustness of the VAE model, which can be improved.}
\label{fig3}
\end{figure}

And similar to the standard deviation, the arithmetic mean and -2/3 mean play roles as measures of the fluctuation.
We will use these metrics to assess the probability inferences. A related generalization of the loss functions, described in the next section, will be used to improve the robustness of the training. Figure \ref{fig2} shows an example of input images from the MNIST dataset and the generated output images produced by the VAE. Despite the blur in some output images, the VAE succeeds in generating very similar images with the input. However,  the histogram in Figure \ref{fig3}, which plots the frequency of the likelihoods over a log scale, shows that the probabilities of ground truth range over a large scale.  The geometric mean or Accuracy captures the central tendency of the distribution at $10^{-37}$ . The Robustness and the Decisiveness capture the span of the fluctuation in the distribution. The -2/3 mean or Robustness is $10^{-77}$ and the arithmetic mean or Decisiveness is $10^{-15}$. The minimal value of -2/3 mean metric is an indicator of the poor robustness of the VAE model, which can be improved.

The metrics derive from a translation of a generalized entropy function back to the probability domain.  Use of the geometric mean for accuracy derives from the Boltzmann-Gibbs-Shannon entropy, which measures the average uncertainty of a system and is equal to the arithmetic average of the negative logarithm of the probability distribution,

\begin{equation}
H\left( \bf{P} \right) \equiv  - \sum\limits_{i = 1}^N {{p_i}\ln {p_i} =  - \ln \left( {\prod\limits_{i = 1}^N {p_i^{{p_i}}} } \right)}.
\end{equation}

Translating the entropy back to the probability domain via the inverse of the negative logarithm, which is the exponential of the negative, results in the weighted geometric mean of the probabilities

\begin{equation}
{P_{avg}} \equiv \exp \left( { - H\left( \bf{P} \right)} \right) = \exp \left( {\ln \left( {\prod\limits_{i = 1}^N {p_i^{{p_i}}} } \right)} \right) = \prod\limits_{i = 1}^N {p_i^{{p_i}}}.
\end{equation}

The role of this function in defining the central tendency of the y-axis of a density is illustrated with the Gaussian distribution.  Utilizing the continuous definition of entropy for a density $f\left( x \right)$ for a random variable $x$, the neutral accuracy or central tendency of the density is

\begin{equation}
{f_{avg}} \equiv \exp \left( { - H\left( {f\left( x \right)} \right)} \right) = \exp \left( {\int\limits_{X} {f\left( x \right)\ln f\left( x \right)} dx} \right).
\end{equation}

For the Gaussian, the average density is equal to the density at the mean plus the standard deviation $f\left( {\mu  \pm \sigma } \right)$.

The use of the geometric mean as a metric for the neutral accuracy in the previous section is related to the cross-entropy between the reported probability of the algorithm and the probability distribution of the test set.  The cross-entropy between a “quoted” or predicted probability distribution $\bf{q}$ and the distribution of the test set $\bf{p}$ is

\begin{equation}\label{cross_entropy}
H\left( {\bf{p},\bf{q}} \right) \equiv  - \sum\limits_i {{p_i}\ln {q_i}}.
\end{equation}

In evaluating an algorithm, the actual distribution is defined by the test samples, which for equally-probable independent samples each have a probability of ${p_i} = \frac{1}{N}$. Translated to the probability domain, the cross-entropy becomes the geometric mean of the reported probabilities \ref{geomean}, thus, showing that use of the geometric mean of the probabilities as a measure of accuracy for reported probabilities is equivalent to the use of cross-entropy as metric of forecasting performance.

Likewise, the use of the generalized mean as a metric from robustness and decisiveness derives from a generalization of the cross-entropy.  While there are a variety of proposed generalizations to information theory, in \citep{Nelson2017, thurner2017three, abe2002stability, renyi1965foundations} the Renyi and Tsallis entropies were both shown to translate to generalized mean upon transformation to the probability domain. Here we show that the derivation of this transformation uses the coupled Entropy, which derives from the Tsallis entropy, but utilizes a modified normalization. The nonlinear statistical coupling  (or simply the coupling), has been shown to a) quantify the relative variance of a superstatistics model in which the variance of exponential distribution fluctuates according to a gamma distribution, and b) be equal to the inverse of the degree of freedom of the Student’s t distribution. The coupling is related to the risk bias by the expression $r = \frac{{ - 2\kappa }}{{1 + \kappa }}$, where the numeral 2 is associated with the power 2 of the Student’s t distribution, and the ratio $r = \frac{{ - 2\kappa }}{{1 + \kappa }}$ is associated with a duality between the positive and negative domains of the coupling. The coupled Entropy uses a generalization of the logarithmic function

\begin{equation}
{\ln _\kappa }\left( x \right) \equiv \frac{1}{\kappa }\left( {{x^\kappa } - 1} \right),\;\,x > 0,
\label{logcoupled}
\end{equation} which provides a continuous set of functions with power. The coupled entropy aggregates the probabilities of a distribution using the generalized mean and translates this to the entropy domain using the generalized logarithm. Using the equiprobable for the sample probabilities, $p_i = \frac{1}{N}$, the coupled cross-entropy “score” for the forecasted probabilities $\bf{q}$ for the event labels $\bf{e}$

\begin{equation}
{S_\kappa }\left( {{\bf{e}},{\bf{q}}} \right) \equiv \frac{{ - 2}}{{1 + \kappa }}{\ln_{\left( {\frac{{ - 2\kappa }}{{1 + \kappa }}} \right)}}\left( {{{\left( {\frac{1}{N}\sum_{i=1}^{N} {q_i^{\frac{{ - 2\kappa }}{{1 + \kappa }}}} } \right)}^{\frac{{ - 1 - \kappa }}{{2\kappa }}}}} \right) \equiv \frac{1}{\kappa }\left( {\left( {\frac{1}{N}\sum_{i=1}^{N} {q_i^{\frac{{ - 2\kappa }}{{1 + \kappa }}}} } \right) - 1} \right),
\end{equation}
where $q_i$ is the probability of event $e_i$ which occurred. Thus the coupled cross-entropy is a local scoring rule dependent only on the probabilities of the actual events.

In order to improve performance against the robust metric, the training of the variational autoencoder needs to incorporate this generalized metric.  To do so we derive a coupled loss function in the next subsection.

\subsection{Coupled loss: Negative ELBO}

As we discussed in the Section 2. The goal of the algorithm is to maximize the variational or evidence lower bound. By defining $L(x^{(i)}) = -ELBO(x^{(i)})$, instead maximizing an Evidence Lower Bound, we minimize its negative $L(x^{(i)})$.


The cross-entropy, which measures the uncertainty of distribution relative to another distribution, underlies both the metrics described above and the loss function used for training. The cross-entropy is the sum of two components, the underlying uncertainty in the distribution $\bf{p}$ measured by the entropy and difference between the distributions measured by the Kullback-Leibler (KL) divergence. The Kullback-Leibler divergence is defined as

\begin{equation}\label{KL_divergence}
{D_{KL}}\left( {\bf{p}||\bf{q}} \right) \equiv  - \sum\limits_i {{p_i}\ln \left( {\frac{{{q_i}}}{{{p_i}}}} \right)}.
\end{equation}

In the VAE algorithm, the loss function consists of the KL-divergence between the posterior approximation $q\left( {{\bf{z}}|{{\bf{x}}^{\left( i \right)}}} \right)$ and a prior $p\left( {\bf{z}} \right)$  and the cross-entropy between the reported probabilities and the training sample distribution.

\begin{equation}
L\left( {{{\bf{x}}^{\left( i \right)}}} \right) =   {D_{KL}}\left( {q\left( {{\bf{z}}|{{\bf{x}}^{\left( i \right)}}} \right)\parallel p\left( {\bf{z}} \right)} \right) - \frac{1}{L}\sum\limits_{l = 1}^L {\left( {\log p\left( {{{\bf{x}}^{\left( i \right)}}|{{\bf{z}}^{\left( {i,l} \right)}}} \right)} \right)},
\end{equation}where the KL-divergence is given by

\begin{equation}
 {D_{KL}}\left( {q\left( {{\bf{z}}|{{\bf{x}}^{\left( i \right)}}} \right)\parallel p\left( {\bf{z}} \right)} \right)= \int\limits_{} q\left( {{\bf{z}}|{{\bf{x}}^{\left( i \right)}}} \right){\left({\log q\left( {{\bf{z}}|{{\bf{x}}^{\left( i \right)}}} \right)}- {\log p(\bf{z})} \right)}d\bf{z}.
\end{equation}

Assume that the posterior distribution of $\bf{x}^{\left(i \right)}$ given $\bf{z}$ is a multivariate Bernoulli, then the cross entropy term is given by

\begin{equation}
    -\frac{1}{L}\sum\limits_{l = 1}^L {\left( {\log p\left( {{{\bf{x}}^{\left( i \right)}}|{{\bf{z}}^{\left( {i,l} \right)}}} \right)} \right)} = -\frac{1}{L}\sum_{l=1}^{L}\sum_{i=1}^{n_x}\Big[x_i\log{y_i}+(1-x_i)\log(1-y_i)\Big],
\end{equation}where $y = \text{Sigmod}(f_2(\tanh{(f_1(\bf{z}))}))$ while $f_1$ and $f_2$ are linear models and $n_x$ is the dimensionality of ${\bf{x}}$.

In this paper, the loss function is modified by coupled generalizations of the KL-divergence and cross-entropy to improve the robustness of the VAE model. The coupled entropy, derived from Nonextensive Statistical Mechanics developed by Tsallis \citep{Tsallis2009}, isolates the degree of nonlinearity which forms the shape of a heavy-tail distribution; the coupling parameter quantifies both the nonlinearity and the tail shape \citep{Nelson2017}. The connection with metrics defined in the previous section is that the generalized loss functions can be derived from the generalized mean. Making use of $r(\kappa,\alpha,d) = \frac{-\alpha \kappa}{1+d\kappa}$ with $\alpha \; = \;2$, the generalized mean is   ${\left( {\sum {p_i^{1 + r}} } \right)^{  \frac{1}{r}} = \left( {\sum {p_i^{1 - \frac{{2\kappa }}{{1 + \kappa }}}} } \right)^{ - \frac{{1 + \kappa }}{{2\kappa }}}}$. When the coupling  $\kappa  \to 0$, the generalized mean is asymptotically equal to the geometric mean.
The coupled entropy function then takes the form of a generalized logarithmic function applied to the generalized mean \citep{Nelson2017}.

\begin{equation}
{H_\kappa }\left( \bf{p} \right) \equiv \frac{1}{2}{\ln _\kappa }\left( {{{\left( {\sum {p_i^{1 + \frac{{2\kappa }}{{1 + \kappa }}}} } \right)}^{\frac{{ - 1}}{\kappa }}}} \right) \equiv \frac{p_i^{1 + \frac{2\kappa }{1 + \kappa }}}{2\sum {p_i^{\frac{{1 + 2\kappa }}{{1 + \kappa }}}}}{\ln _\kappa p_i^{-\frac{2}{1+\kappa}}} \equiv \frac{1}{2\kappa }\left( {{{\left( {\sum {p_i^{\frac{{1 + 3\kappa }}{{1 + \kappa }}}} } \right)}^{ - 1}} - 1} \right),
\end{equation} where  ${\ln _\kappa }\left( x \right)$ is the generalization of the logarithm function in Eq.\ref{logcoupled}.

Similar to the generalization of coupled entropy function, we applied the generalized logarithmic transformation to the original KL-divergence. The first term in KL-divergence becomes 


\begin{equation}
-\int\limits_{} q\left( {{\bf{z}}|{{\bf{x}}^{\left( i \right)}}} \right){\left({\log q\left( {{\bf{z}}|{{\bf{x}}^{\left( i \right)}}} \right)} \right)}d{\bf{z}} \Rightarrow \frac{1}{2}\int_{} \frac{{q{{\left( {{z_j|\bf{x}^{\left( i \right)}}} \right)}^{1 + \frac{{2\kappa }}{{1 + \kappa }}}}}}{{\int\limits_{}  {q{{\left( {{z_j|\bf{x}^{\left( i \right)}}} \right)}^{1 + \frac{{2\kappa }}{{1 + \kappa }}}}d{z_j}} }} {{\ln }_\kappa }\left( {q{{\left( {{z_j|\bf{x}^{\left( i \right)}}} \right)}^{ - \frac{2}{{1 + \kappa }}}}} \right)d{\bf{z}}, 
\end{equation} and the second term in KL-divergence becomes

\begin{equation}
    -\int\limits_{ } q\left( {{\bf{z}}|{{\bf{x}}^{\left( i \right)}}} \right){\left({\log p\left( {{\bf{z}}} \right)} \right)}d{\bf{z}} \Rightarrow \frac{1}{2}\int_{} \frac{{q{{\left( {{z_j|\bf{x}^{\left( i \right)}}} \right)}^{1 + \frac{{2\kappa }}{{1 + \kappa }}}}}}{{\int\limits_{}  {q{{\left( {{z_j|\bf{x}^{\left( i \right)}}} \right)}^{1 + \frac{{2\kappa }}{{1 + \kappa }}}}d{z_j}} }}{{{\ln }_\kappa }\left( {p{{\left( {{z_j}} \right)}^{ - \frac{2}{{1 + \kappa }}}}} \right)}d{\bf{z}}.
\end{equation}

Therefore, the coupled divergence, which is the generalization of KL-divergence in equation \ref{KL_divergence}, with $n_z$ as the dimensionality of ${\bf{z}}$, can be written as

\begin{equation}
    \begin{split}
        &{D_{\kappa}}\left( {q\left( {{\bf{z}}|{{\bf{x}}^{\left( i \right)}}} \right)\parallel p\left( {\bf{z}} \right)} \right)\\
        \equiv & \prod\limits_{j = 1}^{n_z} {\int\limits_{}  {\frac{{q{{\left( {{z_j|\bf{x}^{\left( i \right)}}} \right)}^{1 + \frac{{2\kappa }}{{1 + \kappa }}}}}}{{\int\limits_{}  {q{{\left( {{z_j|\bf{x}^{\left( i \right)}}} \right)}^{1 + \frac{{2\kappa }}{{1 + \kappa }}}}d{z_j}} }}\frac{1}{2}\left({{\ln }_\kappa }\left( {q{{\left( {{z_j|\bf{x}^{\left( i \right)}}} \right)}^{ - \frac{2}{{1 + \kappa }}}}} \right) - {{{\ln }_\kappa }\left( {p{{\left( {{z_j}} \right)}^{ - \frac{2}{{1 + \kappa }}}}} \right)} \right)} d{z_j}} \\
        = & \prod\limits_{j = 1}^{n_z} {\frac{1}{2\kappa }\int\limits_{}  {\frac{{{{\left( {\frac{1}{{\sigma \sqrt {2\pi } }}{e^{ - \frac{{{{\left( {{z_j} - {\mu _i}} \right)}^2}}}{{2{\sigma ^2}}}}}} \right)}^{1 + \frac{{2\kappa }}{{1 + \kappa }}}}}}{{\int\limits_{}  {{{\left( {\frac{1}{{\sigma \sqrt {2\pi } }}{e^{ - \frac{{{{\left( {{z_j} - {\mu _i}} \right)}^2}}}{{2{\sigma ^2}}}}}} \right)}^{1 + \frac{{2\kappa }}{{1 + \kappa }}}}d{z_j}} }} \cdot } \left({{\left( {\frac{1}{{\sigma \sqrt {2\pi } }}{e^{ - \frac{{{{\left( {{z_j} - {\mu _i}} \right)}^2}}}{{2{\sigma ^2}}}}}} \right)}^{ - \frac{{2\kappa }}{{1 + \kappa }}}} - {{{\left( {\frac{1}{{\sqrt {2\pi } }}{e^{ - \frac{{{z_j}^2}}{2}}}} \right)}^{ - \frac{{2\kappa }}{{1 + \kappa }}}}} \right)d{z_j}} 
    \end{split}.
\end{equation}

The original cross entropy can also be modified in a similar way. Applying the genealization of the logarithmic function,  the terms $\log(y_i)$ and $\log(1-y_i)$ are modified to $\frac{1}{2}{{\ln }_\kappa }\left( {{{\left( {{y_i}} \right)}^{\frac{2}{{1 + \kappa }}}}} \right)$  and $\frac{1}{2}{{\ln }_\kappa }\left( {{{\left( {1 - {y_i}} \right)}^{\frac{2}{{1 + \kappa }}}}} \right)$, thus

\begin{equation}
\log p\left( {{{\bf{x}}^{\left( i \right)}}|{{\bf{z}}^{\left( {i,l} \right)}}} \right) \Rightarrow \sum\limits_{i = 1}^{n_x} {\left( {{x_i}\frac{1}{2}{{\ln }_\kappa }\left( {{{\left( {{y_i}} \right)}^{\frac{2}{{1 + \kappa }}}}} \right) + \left( {1 - {x_i}} \right)\frac{1}{2}{{\ln }_\kappa }\left( {{{\left( {1 - {y_i}} \right)}^{\frac{2}{{1 + \kappa }}}}} \right)} \right)}.
\end{equation}

Therefore, the coupled cross-entropy is the generalization of cross-entropy term in equation \ref{cross_entropy}, which is defined as,

\begin{equation}
    H_\kappa ^{\left( l \right)}\left(x_i,y_i\right) \equiv -\frac{1}{2L}\sum_{l=1}^{L}\sum\limits_{i = 1}^{n_x} {\left( {{x_i}{{\ln }_\kappa }\left( {{{\left( {{y_i}} \right)}^{\frac{2}{{1 + \kappa }}}}} \right) + \left( {1 - {x_i}} \right){{\ln }_\kappa }\left( {{{\left( {1 - {y_i}} \right)}^{\frac{2}{{1 + \kappa }}}}} \right)} \right)}.
\end{equation}

The new loss function is the coupled loss function, which is written by

\begin{equation}
L_{\kappa}\left( {{{\bf{x}}^{\left( i \right)}}} \right) =   {D_{\kappa}}\left( {q\left( {{\bf{z}}|{{\bf{x}}^{\left( i \right)}}} \right)\parallel p\left( {\bf{z}} \right)} \right) + H_\kappa ^{\left( l \right)}\left(x_i,y_i\right). 
\end{equation}

\begin{figure}
\centering
        \begin{subfigure}[b]{0.2\textwidth}
                \centering
                \includegraphics[width=.85\linewidth]{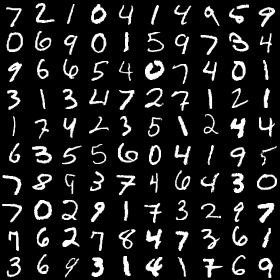}
                \caption{Input Image}
        \end{subfigure}%
                \begin{subfigure}[b]{0.2\textwidth}
                \centering
                \includegraphics[width=.85\linewidth]{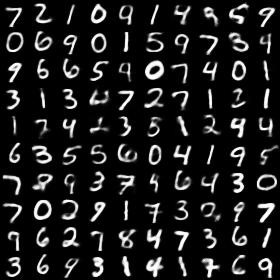}
                \caption{$\kappa = 0$}
        \end{subfigure}%
                \begin{subfigure}[b]{0.2\textwidth}
                \centering
                \includegraphics[width=.85\linewidth]{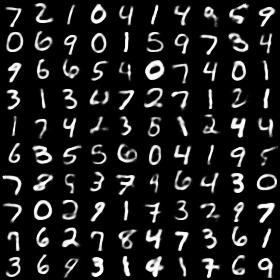}
                \caption{$\kappa = 0.025$}
        \end{subfigure}%
                \begin{subfigure}[b]{0.2\textwidth}
                \centering
                \includegraphics[width=.85\linewidth]{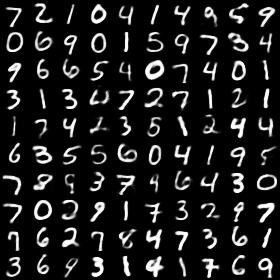}
                \caption{$\kappa = 0.05$}
        \end{subfigure}%
                \begin{subfigure}[b]{0.2\textwidth}
                \centering
                \includegraphics[width=.85\linewidth]{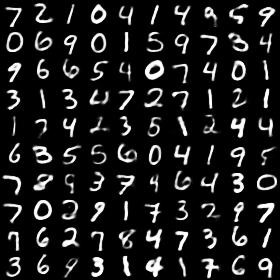}
                \caption{$\kappa = 0.1$}
        \end{subfigure}%
\caption{ (\textbf{a}) The MNIST input images and (\textbf{b}) the output images generated by original VAE. (\textbf{c-e}) The output images generated by modified coupled VAE model show small improvements in detail and clarity. For instance, the fifth digit in the first row of the input images is “4”, but the output image in the original VAE is more like “9” rather than “4” while the coupled VAE method generates “4” correctly.}
\label{fig4}
\end{figure}

Reasons that the coupled loss function can be used to improve the robustness of algorithm include: 1) Higher Uncertainty. The coupled entropy weights low probabilities with a higher cost, forcing the model to increase the probability learned for outliers in the training set.  This ensures that outliers in the test set will be not be over-confident. 2) Penalty for Outliers.
By modeling the correlation between samples, we are discounting the amount of available information. This forces the trained model to have more certainty and thereby be robust against outliers.

\section{Results Using the MNIST Handwritten Numerals}
We trained and tested the coupled VAE model using the MNIST dataset. The algorithm and experiments are developed with Python and the TensorFlow library. Our Python code can be found by Data Availability Statement.

\begin{figure}
\centering
        \begin{subfigure}[b]{0.5\textwidth}
                \centering
                \includegraphics[width=.85\linewidth]{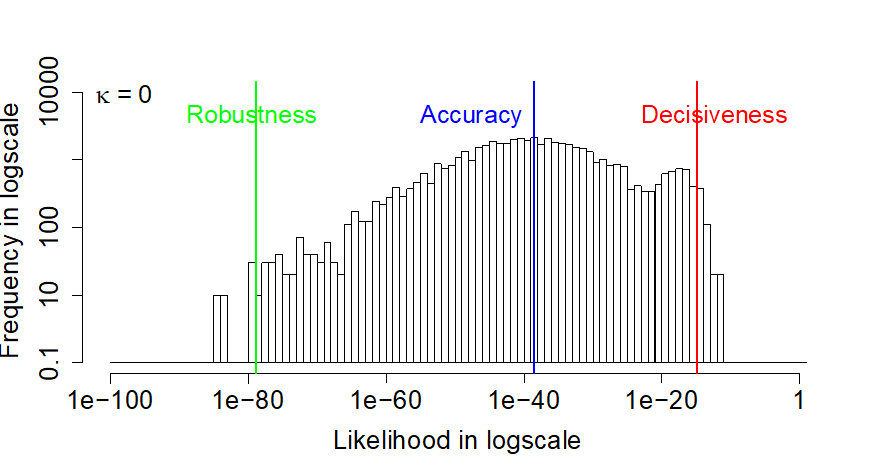}
                \caption{$\kappa = 0$}
        \end{subfigure}%
                \begin{subfigure}[b]{0.5\textwidth}
                \centering
                \includegraphics[width=.85\linewidth]{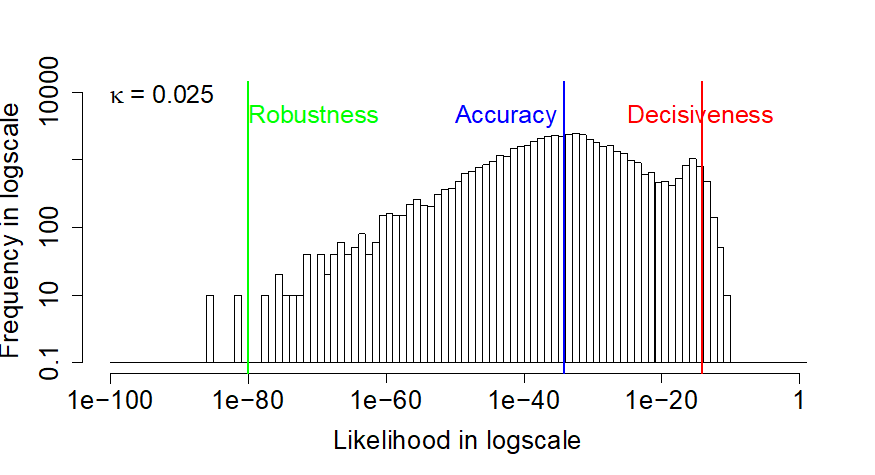}
                \caption{$\kappa = 0.025$}
        \end{subfigure}%
                \vskip\baselineskip

                \begin{subfigure}[b]{0.5\textwidth}
                \centering
                \includegraphics[width=.85\linewidth]{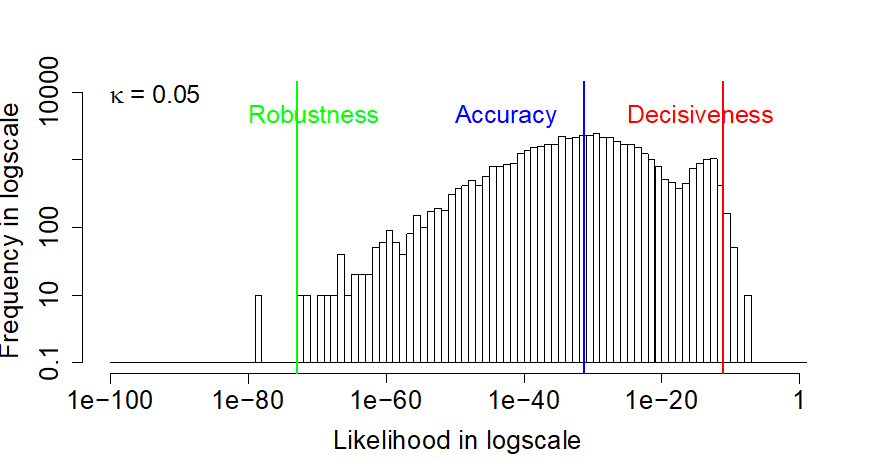}
                \caption{$\kappa = 0.05$}
        \end{subfigure}%
                \begin{subfigure}[b]{0.5\textwidth}
                \centering
                \includegraphics[width=.85\linewidth]{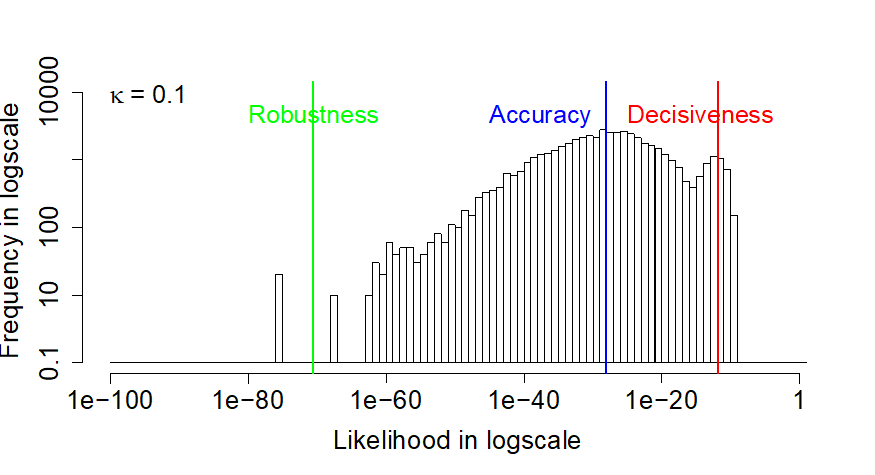}
                \caption{$\kappa = 0.1$}
        \end{subfigure}%

\caption{The histograms of likelihood for the input images with various $\kappa$ values. The red, blue, and green lines represent the arithmetic mean (decisiveness), geometric mean (central tendency), and -2/3 mean (robustness), respectively. The minimal value of the robustness metric indicates that the original VAE suffers from poor robustness. As $\kappa$ gets large, the geometric mean and the -2/3 mean metrics start to increase while the arithmetic mean metric almost keeps same.}
\label{fig5}
\end{figure}

\begin{table} 
\centering
\begin{tabular}{cccc}
\toprule
\textbf{Coupling $\kappa$}	& \textbf{Arithmetic mean metric}	& \textbf{Geometric mean metric}& \textbf{$-\frac{2}{3}$ mean metric}\\
\midrule
 $0$ & $1.31 \times 10^{-15}$ & $2.41 \times 10^{-39}$ & $1.40 \times 10^{-79}$\\
 $0.025$ & $6.61 \times 10^{-15}$ & $5.89 \times 10^{-35}$ & $9.91 \times 10^{-81}$\\
 $0.05$ & $7.18 \times 10^{-12}$ & $5.80 \times 10^{-32}$ & $1.31 \times 10^{-73}$\\
 $0.1$ & $1.34 \times 10^{-12}$ & $7.09 \times 10^{-29}$ & $2.57 \times 10^{-71}$\\
\bottomrule
\end{tabular}
\caption{The relationship between coupling $\kappa$ with the probabilities for input data.\label{tab1}}

\end{table}

The input images and output images for different values of coupling $\kappa$ are shown in Figure \ref{fig4}. $\kappa = 0$  represents the original VAE model. Compared with the original algorithm, output images generated by the modified coupled VAE model show small improvements in detail and clarity. For instance, the fifth digit in the first row of the input images is “4”, but the output image in the original VAE is more like “9” rather than “4” while the coupled VAE method generates “4” correctly. For the seventh digit “4” in the first row, the generated image in the coupled VAE has an improved clarity than the regular VAE.

\begin{figure}
\centering
        \begin{subfigure}[b]{0.25\textwidth}
                \centering
                \includegraphics[width=.85\linewidth]{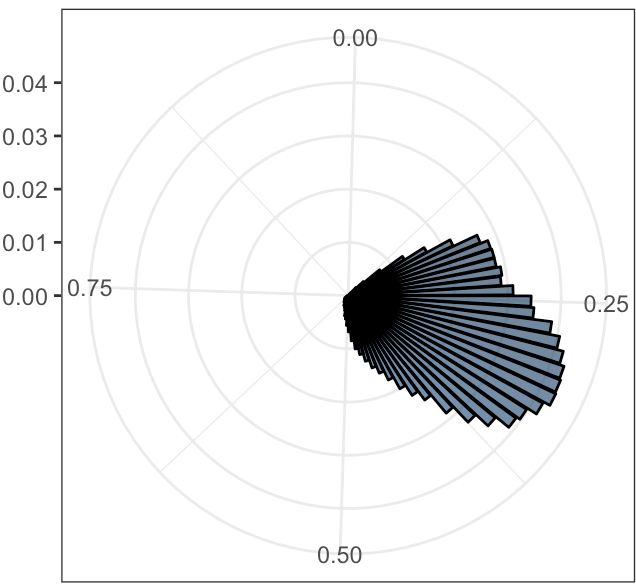}
                \caption{$\kappa = 0$}
        \end{subfigure}%
                \begin{subfigure}[b]{0.25\textwidth}
                \centering
                \includegraphics[width=.85\linewidth]{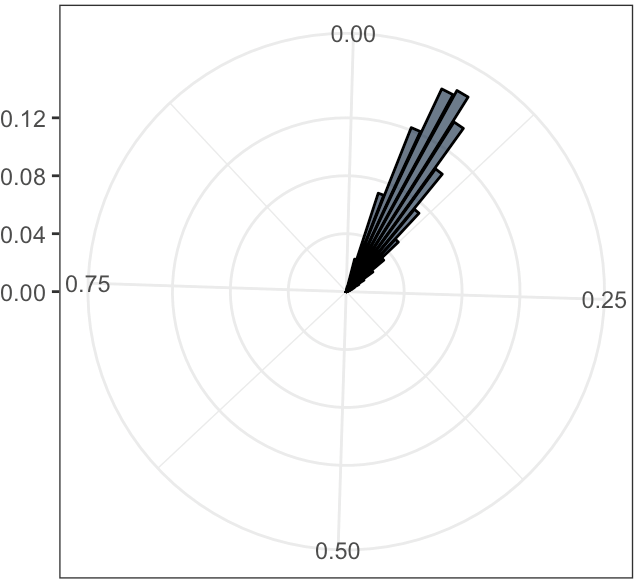}
                \caption{$\kappa = 0.025$}
        \end{subfigure}%
                \begin{subfigure}[b]{0.25\textwidth}
                \centering
                \includegraphics[width=.85\linewidth]{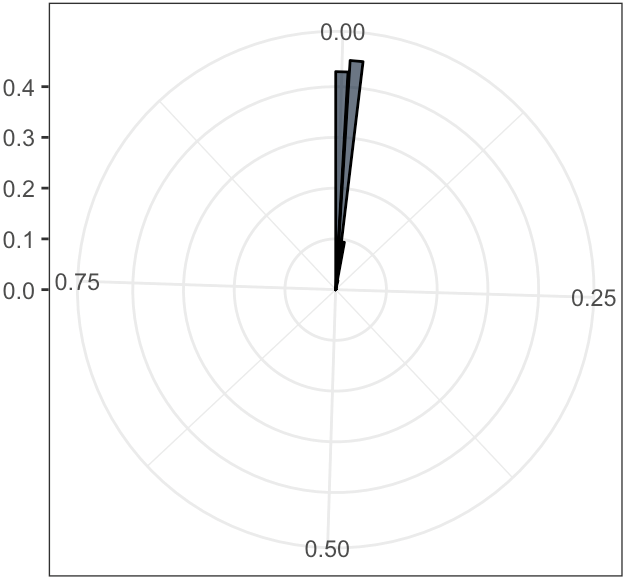}
                \caption{$\kappa = 0.05$}
        \end{subfigure}%
                \begin{subfigure}[b]{0.25\textwidth}
                \centering
                \includegraphics[width=.85\linewidth]{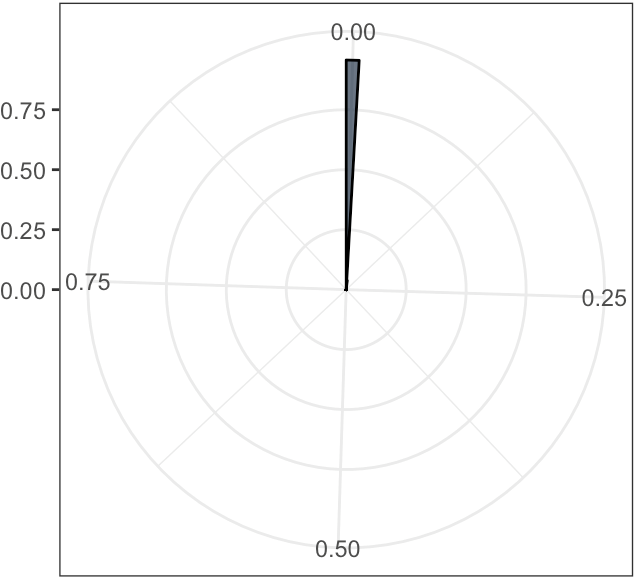}
                \caption{$\kappa = 0.1$}
        \end{subfigure}%
\caption{The rose plots of the various standard deviation values in 20 dimensions. The range and average values of these standard deviations reduce as coupling  increasing.}
\label{fig6}
\end{figure}

\begin{figure}
\centering
        \begin{subfigure}[b]{0.5\textwidth}
                \centering
                \includegraphics[width=.85\linewidth]{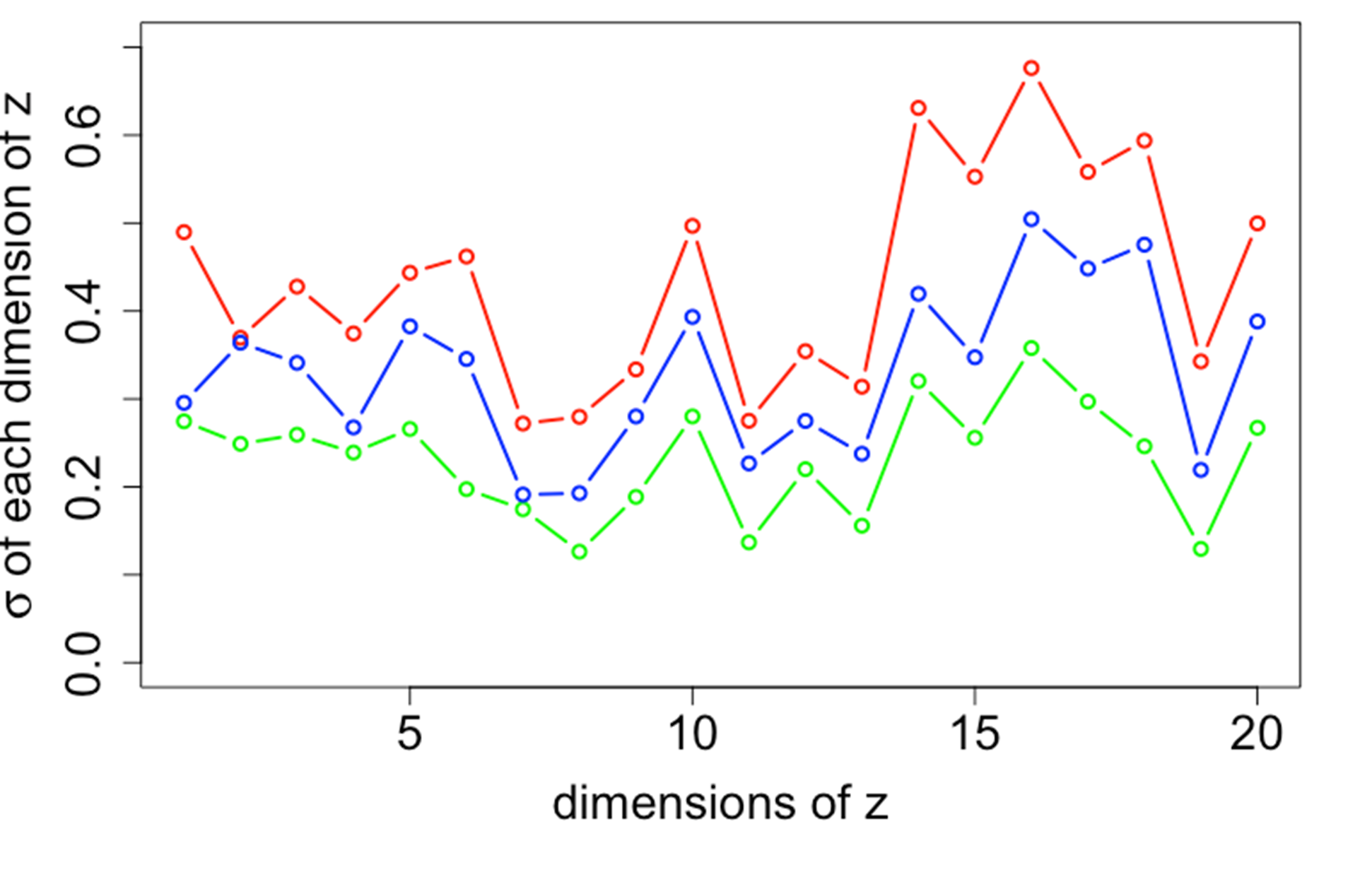}
                \caption{$\kappa = 0$}
        \end{subfigure}%
                \begin{subfigure}[b]{0.5\textwidth}
                \centering
                \includegraphics[width=.85\linewidth]{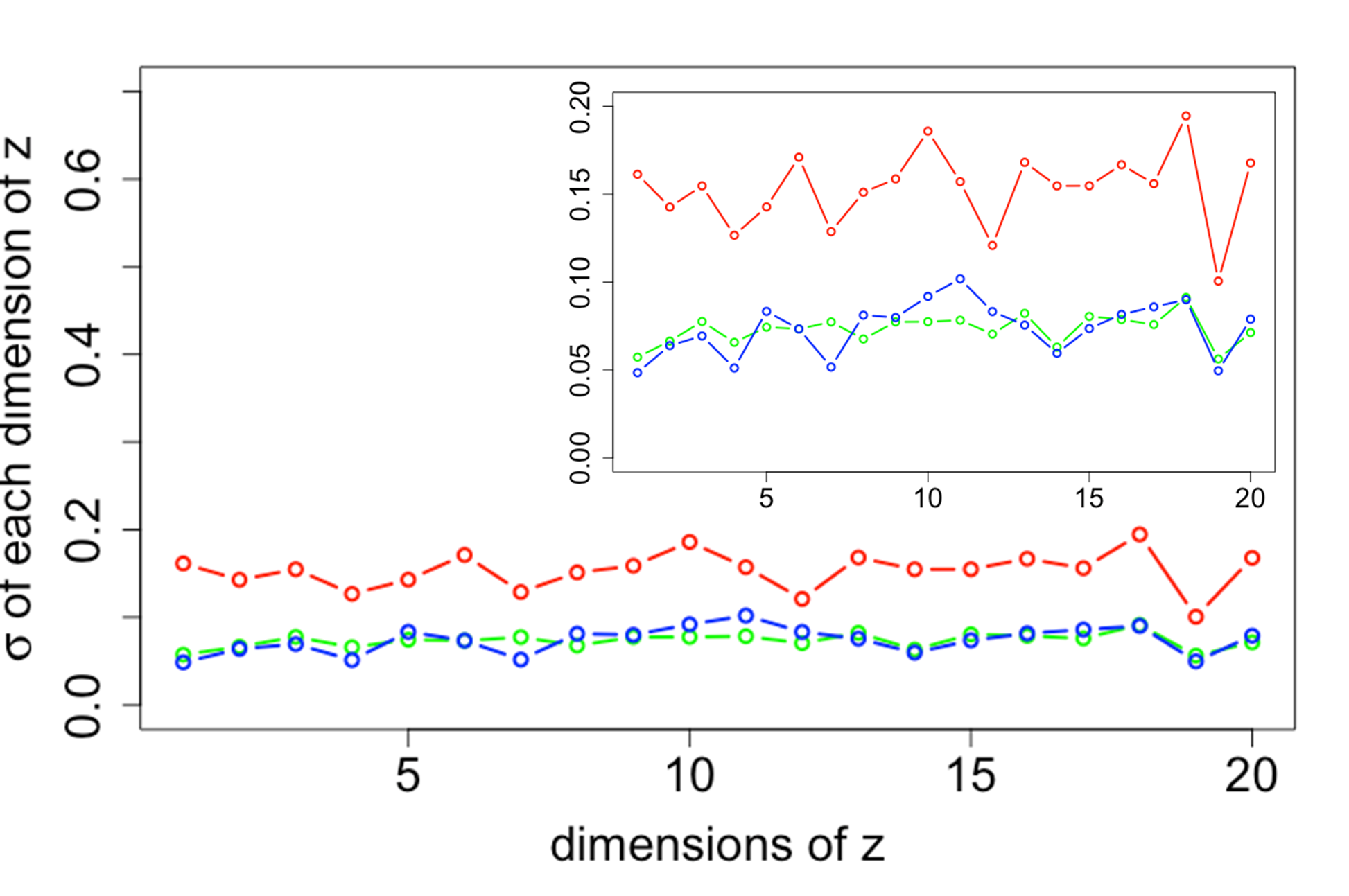}
                \caption{$\kappa = 0.025$}
        \end{subfigure}%
                \vskip\baselineskip
                \begin{subfigure}[b]{0.5\textwidth}
                \centering
                \includegraphics[width=.85\linewidth]{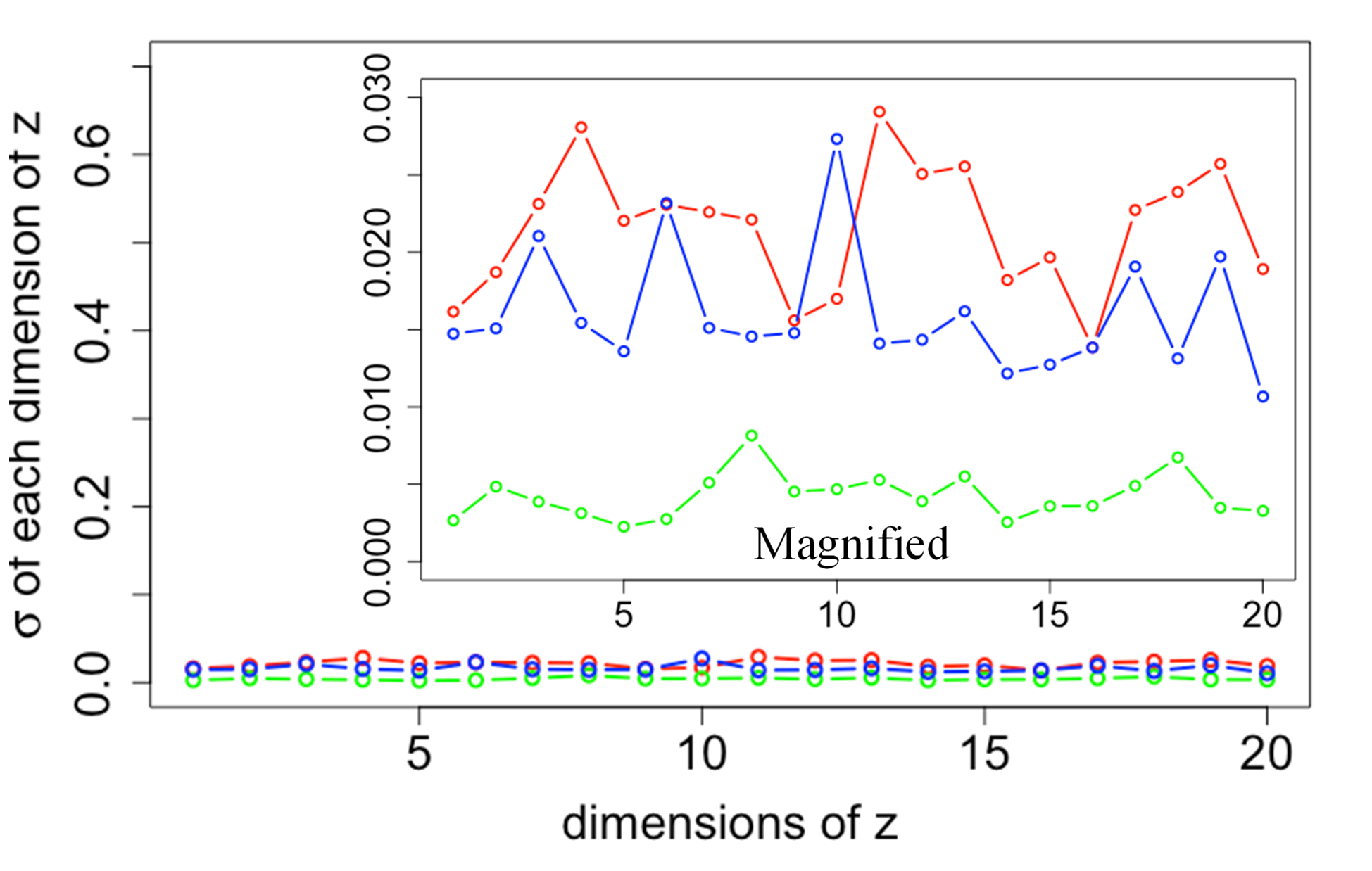}
                \caption{$\kappa = 0.05$}
        \end{subfigure}%
                \begin{subfigure}[b]{0.5\textwidth}
                \centering
                \includegraphics[width=.85\linewidth]{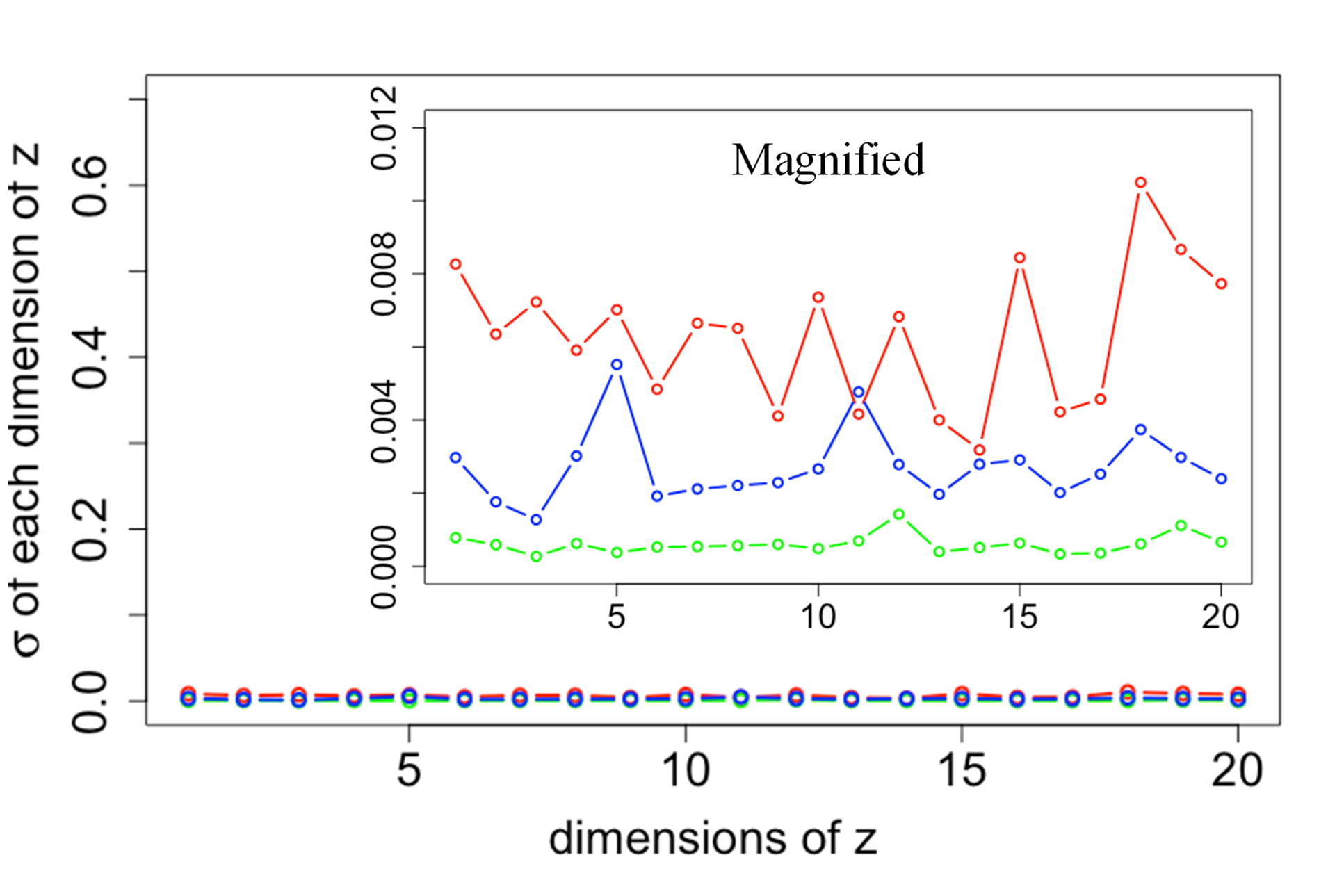}
                \caption{$\kappa = 0.1$}
        \end{subfigure}%
\caption{The standard deviation of latent variable samples near the three generalized mean metrics. The red, blue and green lines represent samples near the decisiveness, accuracy and robustness, respectively. As $\kappa$ increases, values of $\sigma$ are less fluctuant and decrease toward 0. Magnified plots are shown to visualize the results further.}
\label{fig7}
\end{figure}

Figure \ref{fig5} shows the likelihood histograms for 5000 input images with coupling  values of $\kappa = 0, 0.025,$ $ 0.05, 0.1$. The red, blue, and green lines represent the arithmetic mean (decisiveness), geometric mean (central tendency), and -2/3 mean (robustness), respectively. When $\kappa = 0$ , the minimal value of the robustness metric indicates that the original VAE suffers from poor robustness. As $\kappa$ gets large, the geometric mean and the -2/3 mean metrics start to increase while the arithmetic mean metric almost keeps the same. However, when the coupling $\kappa$ becomes large, the coupled loss function can quickly become infinity. For instance, when $\kappa = 0.2$ , the loss function goes infinity at $53^{th}$ epoch; when $\kappa = 0.5$, the loss function goes infinity at $8^{th}$ epoch. In this case, the optimization of coupling values should be further investigated. The specific relationship between coupling $\kappa$ and probabilities for input images is shown in Table \ref{tab1}. The increased robustness metric shows that the modified loss does improve the robustness of the original model.

Furthermore, compared with the original VAE model, the geometric mean, which measures the accuracy of the input image likelihood, is larger for the coupled algorithm. The improvement of this metric means that the input images(truth) are assigned to higher likelihoods in average by the coupled VAE model.

The variance $\sigma$ of latent variables $\bf{z}$ is shown in rose plots in Figure \ref{fig6}. The angular location of a bar represents the value of $\sigma$, clockwise from 0 to 1. The radius of the bar measures the frequency of different $\sigma$ values from 0 to 100. As the coupling $\kappa$ increases, the range and the average value of these standard deviations decrease. To be specific, when $\kappa = 0$, $\sigma$ of all dimensions in all 5000 batches range from 0.09 to 0.72; when $\kappa = 0.025$, $\sigma$ ranges from 0.02 to 0.3; when $\kappa = 0.05$, $\sigma$ ranges from 0.001 to 0.09; when $\kappa = 0.1$, $\sigma$ ranges from  0.00007 to 0.06.

However, here is an issue, as coupling parameter $\kappa$ increases, the variational component of the latent space diminishes. One possible method to address this problem is to use heavy-tail distribution in the latent layer. \citet{chen2020use} and \citet{Nelson2020} used the Student's t as the distribution \cite{takahashi2018student} of the latent layer to incorporate heavy-tail decay.

\begin{figure}
\centering
        \begin{subfigure}[b]{0.5\textwidth}
                \centering
                \includegraphics[width=.85\linewidth]{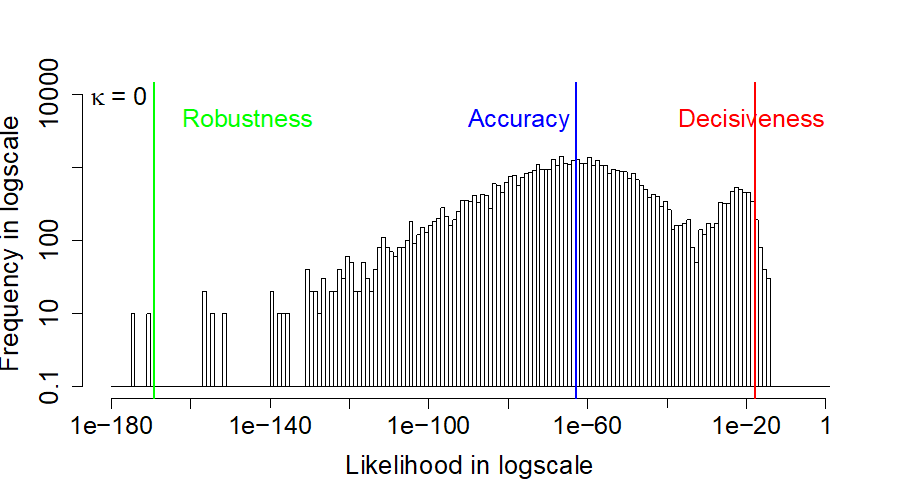}
                \caption{$\kappa = 0$}
        \end{subfigure}%
                \begin{subfigure}[b]{0.5\textwidth}
                \centering
                \includegraphics[width=.85\linewidth]{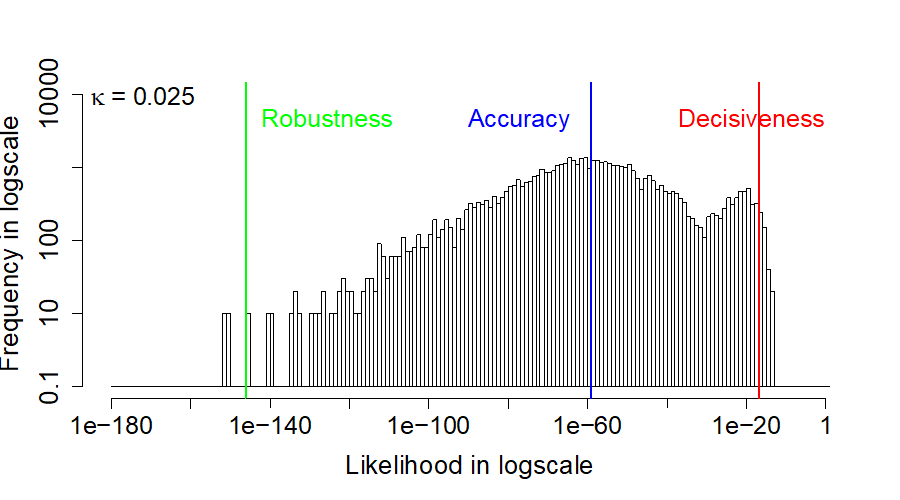}
                \caption{$\kappa = 00.25$}
        \end{subfigure}%
                \vskip\baselineskip
                \begin{subfigure}[b]{0.5\textwidth}
                \centering
                \includegraphics[width=.85\linewidth]{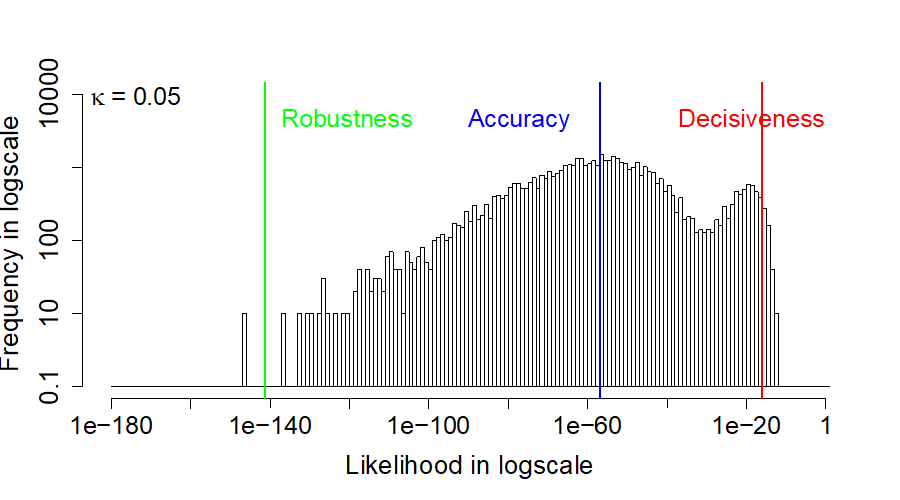}
                \caption{$\kappa = 0.05$}
        \end{subfigure}%
                \begin{subfigure}[b]{0.5\textwidth}
                \centering
                \includegraphics[width=.85\linewidth]{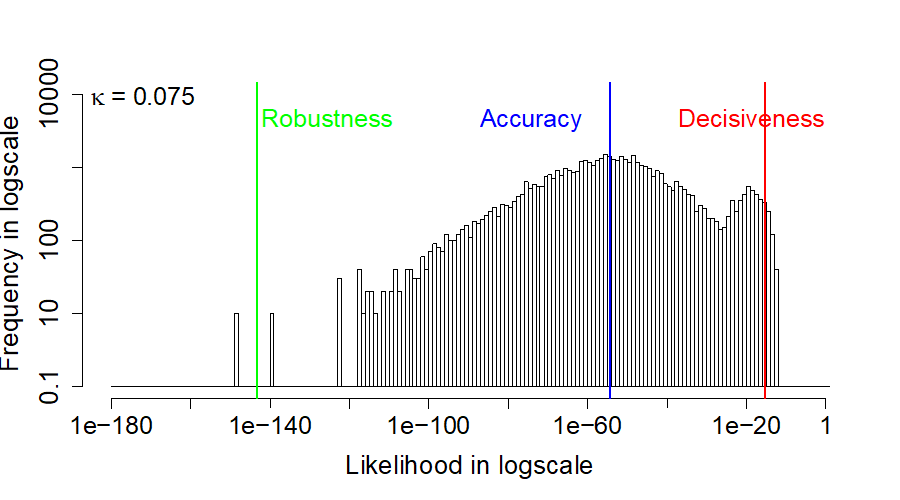}
                \caption{$\kappa = 0.1$}
        \end{subfigure}%
\caption{The histogram likelihood plots with a two-dimensional latent variable. Like the 20-D model, the increased values of arithmetic mean metric and -2/3 mean metric show that the accuracy and robustness of the VAE model have been improved.}
\label{fig8}
\end{figure}

We choose samples in which the likelihoods of input images are close to the three metrics and plotted the standard deviation $\sigma$ of each dimension of the latent variable $\bf{z}$ of these samples in Figure \ref{fig7}. The red, blue and green lines represent samples near the decisiveness, accuracy and robustness, respectively. It shows that when $\kappa = 0$, the standard deviations of $\bf{z}$ range from 0.1 to 0.7. However, as  $\kappa$ increases, values of $\sigma$ are less fluctuate and decrease toward 0. Magnified plots are shown to visualize the results further. The general trend for $\sigma$ is to be more significant for samples near decisiveness, intermediate near the accuracy and smaller for samples near robustness.  An exception is $\kappa = 0.025$, where $\sigma$ overlaps for samples near the robustness and accuracy. 
The histogram likelihood plots with a two-dimensional latent variable is shown in Figure \ref{fig8}. The increased values of arithmetic mean metric and -2/3 mean metric show that the accuracy and robustness of the output MNIST images in VAE model have been improved, consistent with the result in the 20-D model.

\section{Discussion}
In order to understand the relationship between increasing coupling of the loss function with the means and the standard deviations of the Gaussian model, we examine a two-dimensional model which can be visualized.

\begin{figure}
\centering
        \begin{subfigure}[b]{0.25\textwidth}
                \centering
                \includegraphics[width=.85\linewidth]{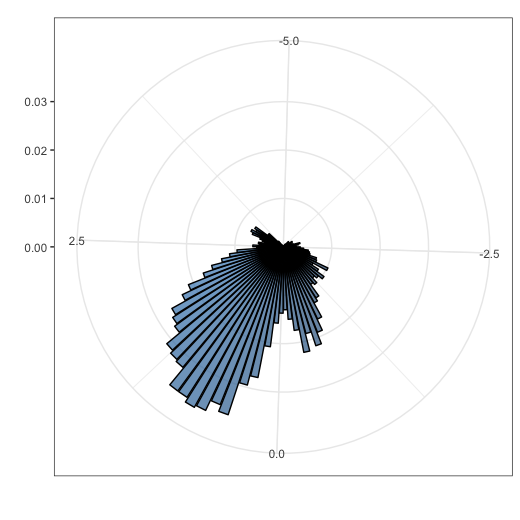}
        \end{subfigure}%
                \begin{subfigure}[b]{0.25\textwidth}
                \centering
                \includegraphics[width=.85\linewidth]{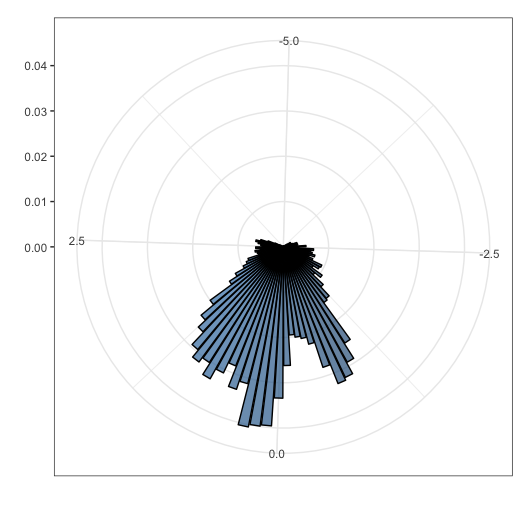}
        \end{subfigure}%
                \begin{subfigure}[b]{0.25\textwidth}
                \centering
                \includegraphics[width=.85\linewidth]{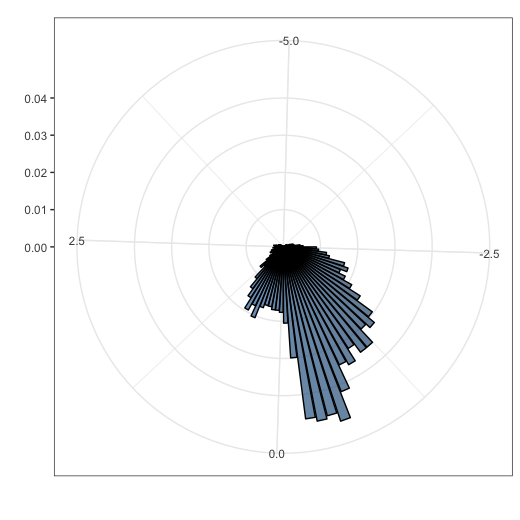}
        \end{subfigure}%
                \begin{subfigure}[b]{0.25\textwidth}
                \centering
                \includegraphics[width=.85\linewidth]{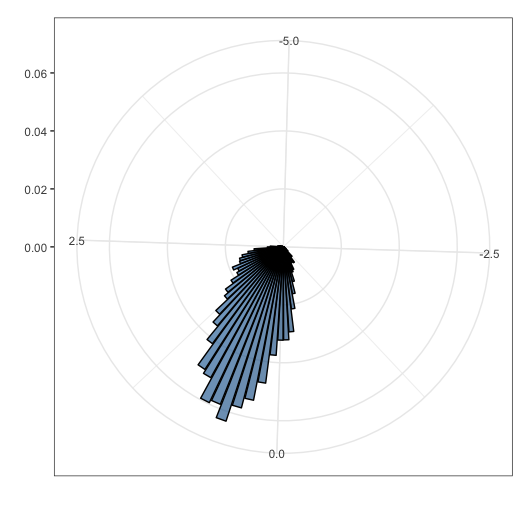}
        \end{subfigure}%
        \\
                \begin{subfigure}[b]{0.25\textwidth}
                \centering
                \includegraphics[width=.85\linewidth]{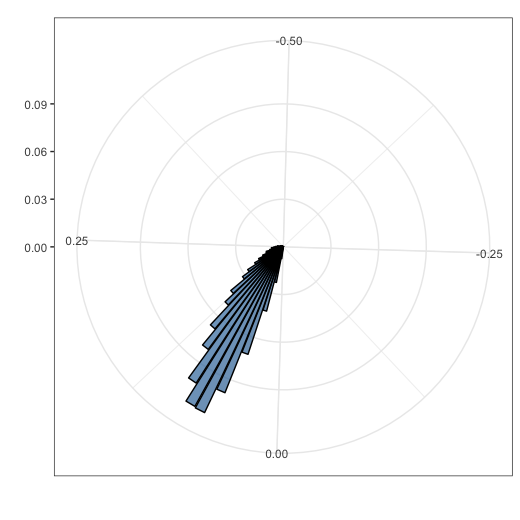}
                \caption{$\kappa = 0$}
        \end{subfigure}%
                \begin{subfigure}[b]{0.25\textwidth}
                \centering
                \includegraphics[width=.85\linewidth]{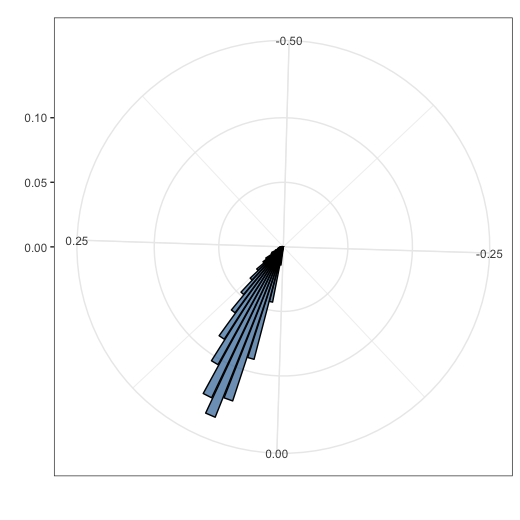}
                \caption{$\kappa = 0.025$}
        \end{subfigure}%
                \begin{subfigure}[b]{0.25\textwidth}
                \centering
                \includegraphics[width=.85\linewidth]{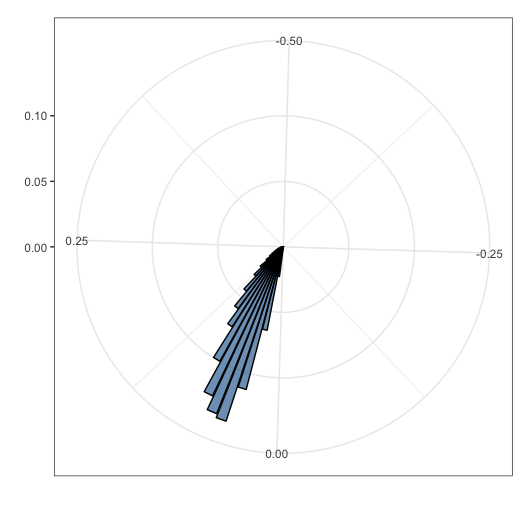}
                \caption{$\kappa = 0.05$}
        \end{subfigure}%
                \begin{subfigure}[b]{0.25\textwidth}
                \centering
                \includegraphics[width=.85\linewidth]{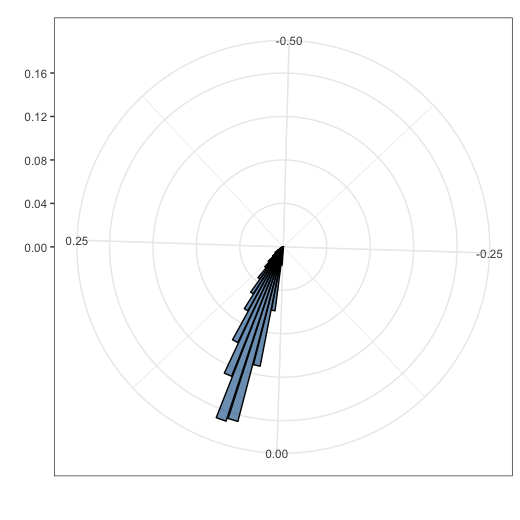}
                \caption{$\kappa = 0.075$}
        \end{subfigure}%
\caption{The rose plots of the various mean (above four figures) and standard deviation (below four figures) values in 2 dimensions.  The range of means reduces and mean values get closer to 0 as coupling increasing.}
\label{fig9}
\end{figure}

Compared with the high-dimensional model, the probability likelihoods for the two-dimensional model are lower, indicating that the higher-dimensions does improve the model. Nevertheless, like the 20-dimensional model, the distribution of likelihood is compressed toward higher values as the coupling increases and, therefore, can be used to analyze the results further. Larger likelihood of input images along with both means closer to the origin and smaller standard deviations of latent variables are the  primary characteristics as the coupling parameter of the loss function is increased. As a result, both the robustness and accuracy metrics of likelihoods increase. To be specific, when $\kappa$ increases from 0 to 0.075, the geometric mean metric increases from $1.20 \times 10^{-63}$ to $4.67 \times 10^{-55}$, and the -2/3 mean metric increases from $5.03 \times 10^{-170}$ to $5.17 \times 10^{-144}$ while the arithmetic metric does not change very much. In this case, the reconstructed images have a higher probability of replicating the input image using the coupled VAE method.

The rose plots in Figure \ref{fig9} shows that the range and variability of the mean values of latent variables decreases as the coupling $\kappa$ increases. From the view of means, the posterior distribution of the latent space is closer to the prior, the standard Gaussian distribution. Moreover, Figure \ref{fig9} show that both the range and the average of the standard deviations decrease to 0 when the coupling $\kappa$ increases. From the view of standard deviations, the posterior distribution of the latent space is further from the prior.

\begin{figure}
\centering
        \begin{subfigure}[b]{0.25\textwidth}
                \centering
                \includegraphics[width=1.2\linewidth]{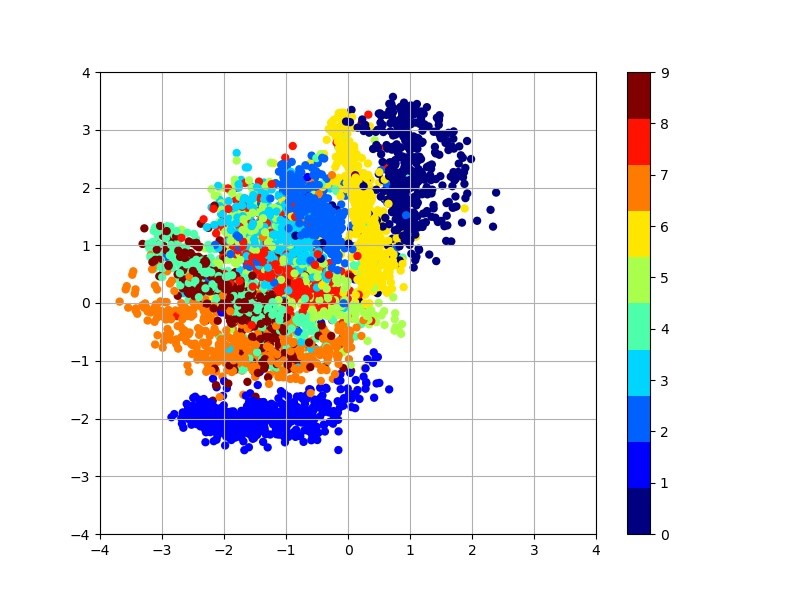}
                \caption{$\kappa = 0$}
        \end{subfigure}%
                \begin{subfigure}[b]{0.25\textwidth}
                \centering
                \includegraphics[width=1.2\linewidth]{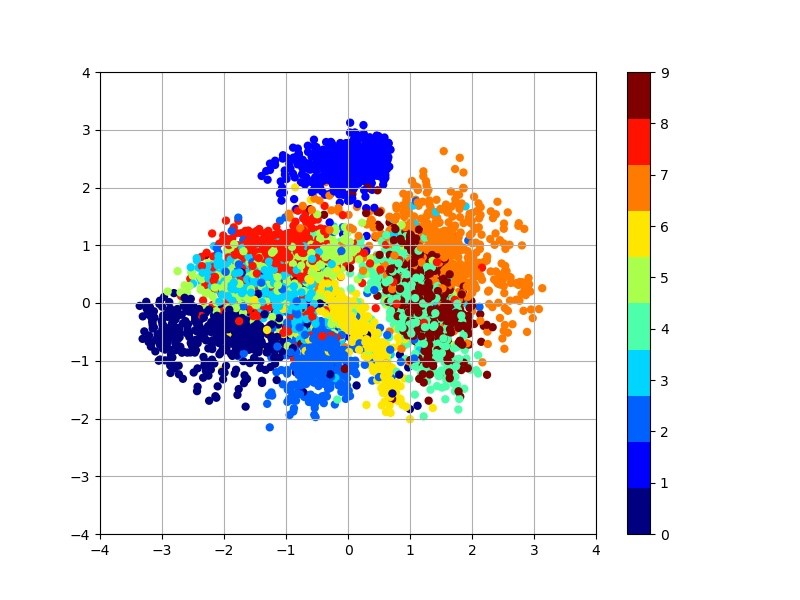}
                \caption{$\kappa = 0.025$}
        \end{subfigure}%
                \begin{subfigure}[b]{0.25\textwidth}
                \centering
                \includegraphics[width=1.2\linewidth]{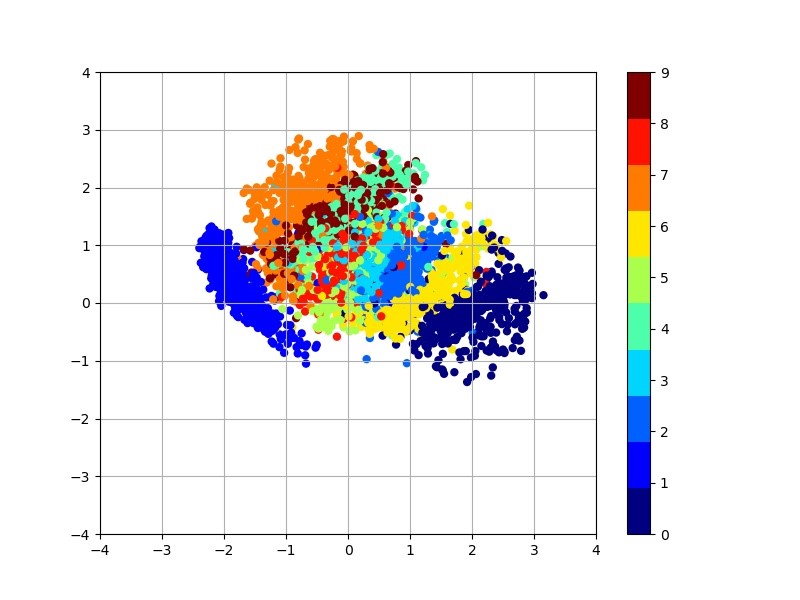}
                \caption{$\kappa = 0.05$}
        \end{subfigure}%
                \begin{subfigure}[b]{0.25\textwidth}
                \centering
                \includegraphics[width=1.2\linewidth]{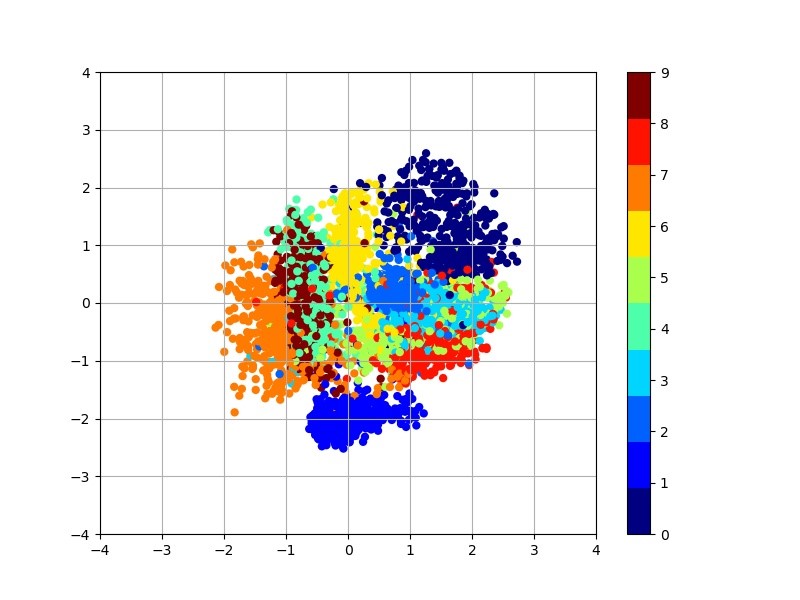}
                \caption{$\kappa = 0.075$}
        \end{subfigure}%
\caption{The plot of the latent space of VAE trained for 200 epochs on MNIST with various $\kappa$ values. Different numerals cluster together more tightly as coupling $\kappa$ increasing.}
\label{fig11}
\end{figure}

\begin{figure}
\centering
        \begin{subfigure}[b]{0.25\textwidth}
                \centering
                \includegraphics[width=.85\linewidth]{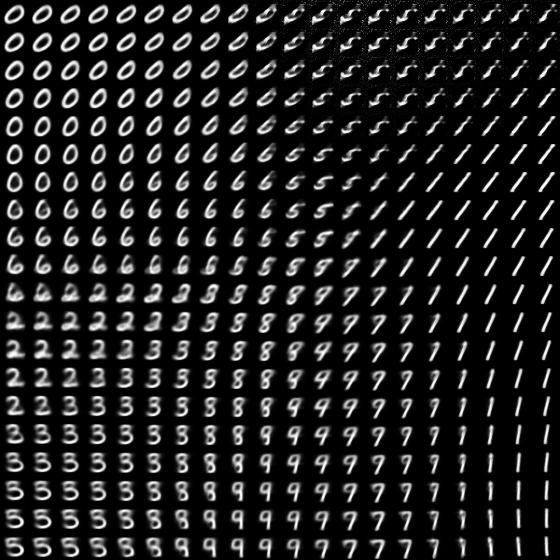}
                \caption{$\kappa = 0$}
        \end{subfigure}%
                \begin{subfigure}[b]{0.25\textwidth}
                \centering
                \includegraphics[width=.85\linewidth]{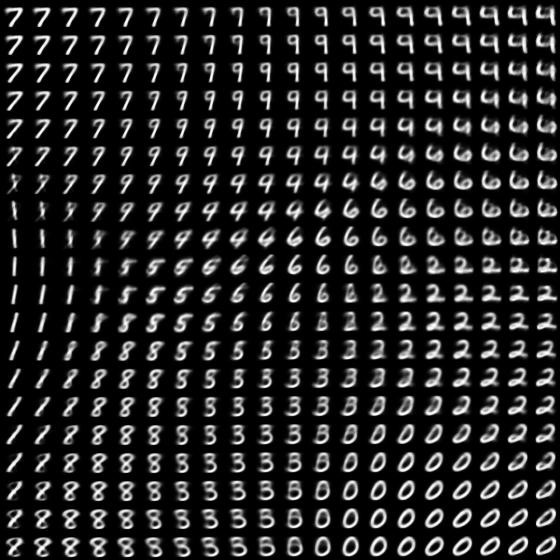}
                \caption{$\kappa = 0.025$}
        \end{subfigure}%
                \begin{subfigure}[b]{0.25\textwidth}
                \centering
                \includegraphics[width=.85\linewidth]{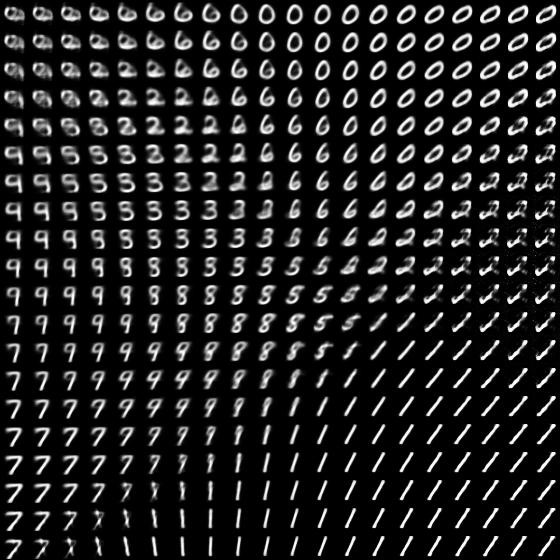}
                \caption{$\kappa = 0.05$}
        \end{subfigure}%
                \begin{subfigure}[b]{0.25\textwidth}
                \centering
                \includegraphics[width=.85\linewidth]{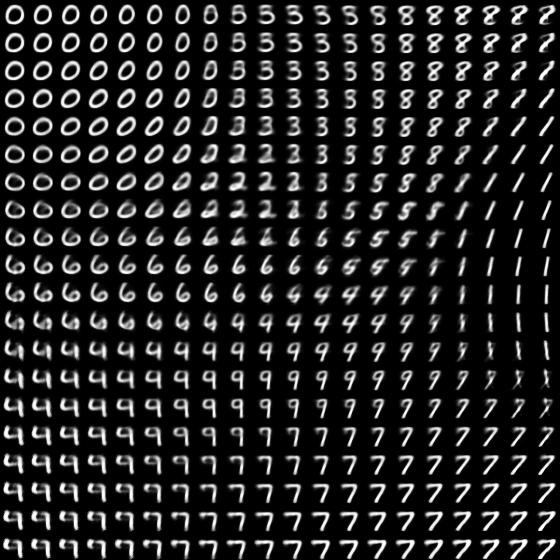}
                \caption{$\kappa = 0.075$}
        \end{subfigure}%
\caption{The plot of visualization of learned data manifold for generative models with the axes to be the values of each dimension of latent variables. The distinct digits each exist in different regions of the latent space and smoothly transform from one digit to another.}
\label{fig12}
\end{figure}

The latent space plots shown in Figure \ref{fig11} are the visualizations of images of the numerals from 0 to 9 Images are embedded in a 2D map where the axis is the values of the 2D latent variable. The same color represents images that belong to the same numeral, and they cluster together since they have higher similarity with each other. The distances between spots represent the similarities of images. The latent space plots show that the different clusters shrink together more tightly when coupling has a large value. The plots shown in Figure \ref{fig12} are the visualization of the learned data manifold generated by the decoder network of the coupled VAE model. A grid of values from a two-dimensional Gaussian distribution is sampled. The distinct digits each exist in different regions of the latent space and smoothly transform from one digit to another. This smooth transformation can be quite useful when the interpolation between two observations is needed. Additionally, the distribution of distinct digits in the plot becomes more even, and the sharpness of the digits increases when $\kappa$ increases.

The reasons why the likelihoods of input images increase and the means and standard deviations decrease of latent variable are analyzed as follows.

\begin{enumerate}
    \item \textit{How do the coupled reconstruction loss and the coupled divergence contribute to total coupled loss function (negative coupled ELBO)?}

    As shown in Table \ref{tab2}, as the coupling increases from 0 to 0.075, the negative ELBO (the loss) decreases from 172.3 to 146.7, the coupled KL-divergence decreases from 5.8 to 5.6, and the coupled reconstruction loss decreases from 166.5 to 141.1. It shows that the reconstruction loss plays a dominant role (with proportion over 96\%), while the divergence term has a much less effect (with proportion under 4\%) in the loss function. The overall improvement of coupled loss is based on both the smaller coupled KL divergence and  the smaller coupled reconstruction error, instead of a trade-off between them. There is a high degree of variability in this improvement, so there ae reasons to be cautious about the degree of improvement. Also, since the coupled loss function is adjusting the metric, the property being measured is also adjusting. Part of our future research plan is to improve how the relative performance between the reconstruction and the latent model can be compared. \\
    
    \item \textit{Why does the latent variable have a smaller deviation in the coupled VAE model?}

    In the coupled VAE algorithm, the loss function is modified to coupled entropy function via the nonlinear statistical coupling. If we consider the values of the latent variable to be locations where an image will be “stored”, then the “nonlinear coupling” models the dependency between these latent values. The coupled VAE method considers long-range correlation among the values of latent variables. If we interpret the dependency to be “similarity”, we can explain the tighter clustering with increased coupling as a result of modeling the dependency. That is because if different latent values, which are representing the images, have more similarities, they will be closer to each other. The shrinkage between numerals corresponds to the decreased variation of the latent variable, thus explaining the smaller standard deviations for the coupled VAE method. \\
    \item \textit{Why do the probabilities of the input images increase in the coupled VAE method?}

    The approximated posterior probability of an input image for the decoder model can be calculated by

    \begin{equation}
    p\left( {x|z} \right) = \frac{{p\left( {x,z} \right)}}{{p\left( z \right)}} = \frac{{p\left( {z|x} \right)p\left( x \right)}}{{p\left( z \right)}} \approx \frac{{q\left( {z|x} \right)p\left( x \right)}}{{p\left( z \right)}}.
    \end{equation}
    following the Gaussian distribution; $p(x)$ is the prior distribution of the input data $x$, $p(z)$ is the prior distribution of latent variable $\bf{z}$, and $q(z|x)$ is the Gaussian approximation of the intractable true posterior $p(z|x)$.

    In our modified algorithm, $p(x)$ and $p(z)$ stay the same, while the Gaussian approximation $q\left( {z|x} \right)$ changes. In the traditional VAE method, we assumed

    \begin{equation}
    {q_1}\left( {z|x} \right) = \frac{1}{{{\sigma _1}\sqrt {2\pi } }}{e^{ - \frac{{{{\left( {z - \mu } \right)}^2}}}{{2\sigma _1^2}}}}.
    \end{equation}
    while our coupled VAE method, assumed

    \begin{equation}
    {q_2}\left( {z|x} \right) = \frac{1}{{{\sigma _2}\sqrt {2\pi } }}{e^{ - \frac{{{{\left( {z - \mu } \right)}^2}}}{{2\sigma _2^2}}}},
    \end{equation} 

    where $\sigma_{2} < \sigma_{1}$ . So, the input data has a smaller range of probabilities, and the average density values increase. In this case, the range of probabilities of input shrinks and the geometric mean of density values increases.
\end{enumerate}

\begin{table}
\centering

\begin{tabular}{cccccc}
\toprule
\textbf{Coupling $\kappa$}	& \textbf{Coupled KL divergence}	& \textbf{- Coupled RE Loss}& \textbf{- Coupled ELBO}& \textbf{KL Proportion}& \textbf{RE Proportion}\\
\midrule
 $0$ & 5.8 +/- 1.7 & 166.5 +/- 52.2 & 172.3 & 3.38\% & 96.62\%\\
 $0.025$ & 5.7 +/- 1.6 & 156.4 +/- 49.8 & 162.1 & 3.53\% & 96.47\%\\
 $0.05$& 5.6 +/- 1.6 & 149.2 +/- 46.6 & 154.8 & 3.61\% & 96.39\%\\
 $0.075$ & 5.6 +/- 1.7 & 141.1 +/- 44.6 & 146.7 & 3.82\% & 96.18\%\\
\bottomrule
\end{tabular}
\caption{Components of ELBO with 2 dimension latent layer under different values of coupling. \label{tab2}}
\end{table}

\section{Conclusions}

This investigation sought to determine whether the accuracy and robustness of Variational Autoencoders can be improved using certain statistical methods developed within the area of complex systems theory. Our investigation provides preliminary evidence that the shape of the negative evidence lower bound loss function can be controlled in such a way that the costs of events in the tails of the loss function are adjustable. We refer to this method as a coupled VAE, since the control parameter models the nonlinear deviation from the exponential and logarithmic functions of linear analysis. A positive coupling parameter increases the cost of these tail events and thereby trains the algorithm to be robust against such outliers. Additionally, this improves both the accuracy of reconstructed images and reduces divergence of posterior latent distributions from priors. We have been able to document this improvement using the histogram of likelihoods for input data based on arithmetic mean, geometric mean, and -2/3 mean, which represent decisiveness, accuracy, and robustness, respectively. Both the accuracy and the robustness are improved by increasing the coupling of the loss function. There is a limit to such increases in the coupling beyond which the training process no longer converges. These performance improvements have been shown for the MNIST handwritten numeral dataset using a 20-dimensional latent layer representation.

The modifications of the latent posterior distributions have been further examined using a two-dimensional representation. We show that the latent variables have both a tighter distribution of the mean about its prior value of zero, and a movement of standard deviations towards zero, away from the prior of one, as coupling $\kappa$ increases. Overall, coupled KL divergence does indeed decrease as the coupling is increased, indicating improvement in the latent representation. Thus improvements in the reconstruction evident from both visual clarity of images and increased accuracy in measured likelihoods is not due to a trade-off with the latent representation. Rather the negative coupled ELBO metric shows improvement in both latent layer divergence and output image reconstruction.

In future research we plan to study the coupled Gaussian distribution as the prior and posterior distribution of the latent layer. This may be helpful for achieving greater separation between the images into distinct clusters similar to what has been achieved with t-Stochastic Neighborhood Embedding methods \citep{VanDerMaaten2008}. If so, it may be possible to improve the decisiveness of the likelihoods in addition to further improvements in the accuracy and robustness. We also plan to further improve the analysis of the latent layer representation, so that this can be examined with higher-dimensional models. Finally, we are planning to expand the evaluation to more complex datasets such as CIFAR-10 which will utilize convolutional neural networks.

\vspace{6pt} 








\bibliographystyle{unsrtnat}
\bibliography{references}  

\begin{thebibliography}{32}
\providecommand{\natexlab}[1]{#1}
\providecommand{\url}[1]{\texttt{#1}}
\expandafter\ifx\csname urlstyle\endcsname\relax
  \providecommand{\doi}[1]{doi: #1}\else
  \providecommand{\doi}{doi: \begingroup \urlstyle{rm}\Url}\fi

\bibitem[Kingma and Welling(2014)]{Kingma2014}
Diederik~P Kingma and Max Welling.
\newblock Auto-encoding variational bayes.
\newblock In \emph{International Conference on Learning Representations (ICLR),
  arXiv: 1312.6114v10}, 2014.
\newblock URL \url{https://arxiv.org/pdf/1312.6114.pdf}.

\bibitem[Tran et~al.(2017)Tran, Hoffman, Saurous, Brevdo, Murphy, and
  Blei]{Tran2017}
Dustin Tran, Matthew~D Hoffman, Rif~A. Saurous, Eugene Brevdo, Kevin Murphy,
  and David~M Blei.
\newblock Deep probabilistic programming.
\newblock In \emph{Fifth International Conference on Learning Representations,
  arXiv:1701.03757}, 2017.
\newblock URL \url{https://arxiv.org/abs/1701.03757}.

\bibitem[Bowman et~al.(2015)Bowman, Vilnis, Vinyals, Dai, Jozefowicz, and
  Bengio]{Samuel2015}
Samuel~R. Bowman, Luke Vilnis, Oriol Vinyals, Andrew~M. Dai, Rafal Jozefowicz,
  and Samy Bengio.
\newblock Generating sentences from a continuous space.
\newblock In \emph{Proceedings of the Twentieth Conference on Computational
  Natural Language Learning (CoNLL), arXiv:1511.06349}, 2015.
\newblock URL \url{http://arxiv.org/abs/1511.06349}.

\bibitem[Zalger(2017)]{Zalger2017}
Jonathan Zalger.
\newblock Application of variational autoencoders for aircraft turbomachinery
  design.
\newblock Technical report, Stanford Univ., 2017.
\newblock URL
  \url{http://cs229.stanford.edu/proj2017/final-reports/5231979.pdf}.

\bibitem[Xu et~al.(2018)Xu, Feng, Chen, Wang, Qiao, Chen, Zhao, Li, Bu, Li,
  Liu, Zhao, and Pei]{Xu2018}
Haowen Xu, Yang Feng, Jie Chen, Zhaogang Wang, Honglin Qiao, Wenxiao Chen,
  Nengwen Zhao, Zeyan Li, Jiahao Bu, Zhihan Li, Ying Liu, Youjian Zhao, and Dan
  Pei.
\newblock Unsupervised anomaly detection via variational auto-encoder for
  seasonal kpis in web applications.
\newblock In \emph{Proceedings of the 2018 World Wide Web Conference on World
  Wide Web. arXiv:1802.03903}, pages 187--196, New York, New York, USA, 2018.
  ACM Press.
\newblock ISBN 9781450356398.
\newblock URL \url{http://dl.acm.org/citation.cfm?doid=3178876.3185996}.

\bibitem[Blei et~al.(2016)Blei, Kucukelbir, and McAuliffe]{Blei2016}
David~M. Blei, Alp Kucukelbir, and Jon~D. McAuliffe.
\newblock {Variational Inference: A Review for Statisticians}.
\newblock \emph{Journal of the American statistical Association}, 112:\penalty0
  859--877, Jan 2016.
\newblock URL \url{http://arxiv.org/abs/1601.00670}.

\bibitem[Higgins et~al.(2016)Higgins, Matthey, Pal, Burgess, Glorot, Botvinick,
  Mohamed, and Lerchner]{higgins2016beta}
Irina Higgins, Loic Matthey, Arka Pal, Christopher Burgess, Xavier Glorot,
  Matthew Botvinick, Shakir Mohamed, and Alexander Lerchner.
\newblock beta-vae: Learning basic visual concepts with a constrained
  variational framework.
\newblock 2016.

\bibitem[Burgess et~al.(2018)Burgess, Higgins, Pal, Matthey, Watters,
  Desjardins, and Lerchner]{burgess2018understanding}
Christopher~P Burgess, Irina Higgins, Arka Pal, Loic Matthey, Nick Watters,
  Guillaume Desjardins, and Alexander Lerchner.
\newblock Understanding disentangling in $beta $-vae.
\newblock \emph{arXiv preprint arXiv:1804.03599}, 2018.

\bibitem[Nelson and Umarov(2010)]{Nelson2010}
Kenric~P. Nelson and Sabir Umarov.
\newblock Nonlinear statistical coupling.
\newblock \emph{Physica A: Statistical Mechanics and its Applications},
  389\penalty0 (11):\penalty0 2157--2163, 2010.
\newblock ISSN 03784371.

\bibitem[Nelson et~al.(2017)Nelson, Umarov, and Kon]{Nelson2017}
Kenric~P. Nelson, Sabir~R. Umarov, and Mark~A. Kon.
\newblock On the average uncertainty for systems with nonlinear coupling.
\newblock \emph{Physica A: Statistical Mechanics and its Applications},
  468:\penalty0 30--43, Feb 2017.
\newblock ISSN 03784371.

\bibitem[Nelson(2020)]{Nelson2020}
Kenric~P. Nelson.
\newblock Reduced perplexity: A simplified perspective on assessing
  probabilistic forecasts.
\newblock In Min Chen, Jon~M. Dunn, Amos Golan, and Aman Ullah, editors,
  \emph{Info-Metrics Volume}. Oxford University Press, 2020.
\newblock URL \url{http://arxiv.org/abs/1603.08830}.

\bibitem[Tsallis(2009)]{Tsallis2009}
Constantino Tsallis.
\newblock \emph{Introduction to nonextensive statistical mechanics: Approaching
  a complex world}.
\newblock Springer New York, 2009.
\newblock ISBN 9780387853581.

\bibitem[Akrami et~al.(2019)Akrami, Joshi, Li, Aydore, and
  Leahy]{akrami2019robust}
Haleh Akrami, Anand~A Joshi, Jian Li, Sergul Aydore, and Richard~M Leahy.
\newblock Robust variational autoencoder.
\newblock \emph{arXiv preprint arXiv:1905.09961}, 2019.

\bibitem[Niemitalo(2010)]{Olli2010}
Olli Niemitalo.
\newblock A method for training artificial neural networks to generate missing
  data within a variable context, 2010.
\newblock URL
  \url{https://web.archive.org/web/20120312111546/http://yehar.com:80/blog/?p=167}.

\bibitem[Goodfellow et~al.(2014)Goodfellow, Pouget-Abadie, Mirza, Xu,
  Warde-Farley, Ozair, Courville, and Bengio]{Goodfellow2014}
Ian Goodfellow, Jean Pouget-Abadie, Mehdi Mirza, Bing Xu, David Warde-Farley,
  Sherjil Ozair, Aaron Courville, and Yoshua Bengio.
\newblock Generative adversarial nets.
\newblock In Z.~Ghahramani, M.~Welling, C.~Cortes, N.~D. Lawrence, and K.~Q.
  Weinberger, editors, \emph{Advances in Neural Information Processing Systems
  27}, pages 2672--2680. Curran Associates, Inc., 2014.
\newblock URL
  \url{http://papers.nips.cc/paper/5423-generative-adversarial-nets.pdf}.

\bibitem[Donahue et~al.(2017)Donahue, Darrell, and
  Kr{\"{a}}henb{\"{u}}hl]{Donahue2017}
Jeff Donahue, Trevor Darrell, and Philipp Kr{\"{a}}henb{\"{u}}hl.
\newblock Adversarial feature learning.
\newblock In \emph{5th International Conference on Learning Representations,
  ICLR 2017 - Conference Track Proceedings}. International Conference on
  Learning Representations, ICLR, 2017.

\bibitem[Dumoulin et~al.(2017)Dumoulin, Belghazi, Poole, Mastropietro, Lamb,
  Arjovsky, and Courville]{Dumoulin2017}
Vincent Dumoulin, Ishmael Belghazi, Ben Poole, Olivier Mastropietro, Alex Lamb,
  Martin Arjovsky, and Aaron Courville.
\newblock Adversarially learned inference.
\newblock In \emph{5th International Conference on Learning Representations,
  ICLR 2017 - Conference Track Proceedings}. International Conference on
  Learning Representations, ICLR, 2017.

\bibitem[Neyshabur et~al.(2017)Neyshabur, Bhojanapalli, and
  Chakrabarti]{neyshabur2017stabilizing}
Behnam Neyshabur, Srinadh Bhojanapalli, and Ayan Chakrabarti.
\newblock Stabilizing gan training with multiple random projections.
\newblock \emph{arXiv preprint arXiv:1705.07831}, 2017.

\bibitem[Pearl(1985)]{Pearl1985}
Judea Pearl.
\newblock Bayesian netwcrks: A model cf self-activated memory for evidential
  reasoning.
\newblock Technical report, University of California, 1985.
\newblock URL \url{http://ftp.cs.ucla.edu/pub/stat{\_}ser/r43-1985.pdf}.

\bibitem[Goodfellow et~al.(2016)Goodfellow, Bengio, and
  Courville]{Goodfellow2016}
Ian Goodfellow, Yoshua Bengio, and Aaron Courville.
\newblock \emph{Deep learning}.
\newblock MIT Press, 2016.
\newblock URL \url{http://www.deeplearningbook.org/}.

\bibitem[Ebbers et~al.(2017)Ebbers, Heymann, Drude, Glarner, Haeb-Umbach, and
  Raj]{Ebbers2017}
Janek Ebbers, Jahn Heymann, Lukas Drude, Thomas Glarner, Reinhold Haeb-Umbach,
  and Bhiksha Raj.
\newblock Hidden markov model variational autoencoder for acoustic unit
  discovery.
\newblock \emph{Interspeech}, 2017.
\newblock URL \url{http://dx.doi.org/10.21437/Interspeech.2017-1160}.

\bibitem[Dilokthanakul et~al.(2016)Dilokthanakul, Mediano, Garnelo, Lee,
  Salimbeni, Arulkumaran, and Shanahan]{Dilokthanakul2016}
Nat Dilokthanakul, Pedro A.~M. Mediano, Marta Garnelo, Matthew C.~H. Lee, Hugh
  Salimbeni, Kai Arulkumaran, and Murray Shanahan.
\newblock Deep unsupervised clustering with gaussian mixture variational
  autoencoders, Nov 2016.
\newblock URL \url{http://arxiv.org/abs/1611.02648}.

\bibitem[Srivastava and Sutton(2017)]{Srivastava2017}
Akash Srivastava and Charles Sutton.
\newblock Autoencoding variational inference for topic models.
\newblock \emph{ICLR 2017}, Mar 2017.
\newblock URL \url{http://arxiv.org/abs/1703.01488}.

\bibitem[LeCun et~al.(1998)LeCun, Cortes, and Burges]{Yann}
Yann LeCun, Corinna Cortes, and Christopher~J.C. Burges.
\newblock The {MNIST} database of handwritten digits, 1998.
\newblock URL \url{http://yann.lecun.com/exdb/mnist/index.html}.

\bibitem[McAlister(1879)]{Donald1879}
Donald McAlister.
\newblock Xiii. the law of the geometric mean.
\newblock \emph{Proceedings of the Royal Society}, 29\penalty0
  (196-199):\penalty0 367--376, Dec 1879.
\newblock ISSN 0370-1662.

\bibitem[Frogner et~al.(2015)Frogner, Zhang, Mobahi, Araya, and
  Poggio]{Frogner2015}
Charlie Frogner, Chiyuan Zhang, Hossein Mobahi, Mauricio Araya, and Tomaso~A
  Poggio.
\newblock Learning with a wasserstein loss.
\newblock In C.~Cortes, N.~D. Lawrence, D.~D. Lee, M.~Sugiyama, and R.~Garnett,
  editors, \emph{{Advances in Neural Information Processing Systems 28}}, pages
  2053--2061. Curran Associates, Inc., 2015.
\newblock URL
  \url{http://papers.nips.cc/paper/5679-learning-with-a-wasserstein-loss.pdf}.

\bibitem[Thurner et~al.(2017)Thurner, Corominas-Murtra, and
  Hanel]{thurner2017three}
Stefan Thurner, Bernat Corominas-Murtra, and Rudolf Hanel.
\newblock Three faces of entropy for complex systems: Information,
  thermodynamics, and the maximum entropy principle.
\newblock \emph{Physical Review E}, 96\penalty0 (3):\penalty0 032124, 2017.

\bibitem[Abe(2002)]{abe2002stability}
Sumiyoshi Abe.
\newblock Stability of tsallis entropy and instabilities of r{\'e}nyi and
  normalized tsallis entropies: A basis for q-exponential distributions.
\newblock \emph{Physical Review E}, 66\penalty0 (4):\penalty0 046134, 2002.

\bibitem[R{\'e}nyi(1965)]{renyi1965foundations}
Alfr{\'e}d R{\'e}nyi.
\newblock On the foundations of information theory.
\newblock \emph{Revue de l'Institut International de Statistique}, pages 1--14,
  1965.

\bibitem[Chen et~al.(2020)Chen, Svoboda, and Nelson]{chen2020use}
Kevin~R Chen, Daniel Svoboda, and Kenric~P Nelson.
\newblock Use of student's t-distribution for the latent layer in a coupled
  variational autoencoder.
\newblock \emph{arXiv preprint arXiv:2011.10879}, 2020.

\bibitem[Takahashi et~al.(2018)Takahashi, Iwata, Yamanaka, Yamada, and
  Yagi]{takahashi2018student}
Hiroshi Takahashi, Tomoharu Iwata, Yuki Yamanaka, Masanori Yamada, and Satoshi
  Yagi.
\newblock Student-t variational autoencoder for robust density estimation.
\newblock In \emph{IJCAI}, pages 2696--2702, 2018.

\bibitem[{Van Der Maaten} and Hinton(2008)]{VanDerMaaten2008}
Laurens {Van Der Maaten} and Geoffrey Hinton.
\newblock Visualizing data using {T-SNE}.
\newblock \emph{Journal of Machine Learning Research}, 9:\penalty0 2579--2605,
  2008.

\end{thebibliography}






\end{document}